\documentclass[letterpaper, 10 pt, journal, twoside]{IEEEtran} 

\pdfoutput=1
\usepackage{graphics} 
\usepackage{subcaption}
\usepackage{multirow}
\usepackage[hyphens]{url}
\usepackage{hyperref}
\usepackage{cite}
\usepackage{epsfig} 
\usepackage{mathptmx} 
\usepackage{times} 
\usepackage{amsmath,amssymb,amsfonts,mathrsfs} 
\usepackage{txfonts} 
\usepackage{algorithm,algorithmic}

\usepackage{todonotes}
\usepackage{soul}
\definecolor{smoothgreen}{rgb}{0.7,1,0.7}
\sethlcolor{smoothgreen}

\begin{document}

\title{Interactive Natural Language-based Person Search}

\author{Vikram Shree$^{1}$, Wei-Lun Chao$^{2}$, and Mark Campbell$^{1}$%
\thanks{$^{1}$Vikram Shree and Mark Campbell are with the Sibley School of Mechanical and Aerospace Engineering, Cornell University, USA
        {\tt\footnotesize \{vs476, mc288\}@cornell.edu}}%
\thanks{$^{2}$Wei-Lun Chao is with the Department of Computer Science and Engineering, The Ohio State University, USA
        {\tt\footnotesize chao.209@osu.edu}}%
\thanks{\textbf{Cite as:} V. Shree, W. Chao, and M. Campbell. ``Interactive Natural Language-based Person Search." in IEEE Robotics and Automation Letters, vol. 5, no. 2, pp. 1851-1858, April 2020.}
\thanks{\textbf{Digital Object Identifier (DOI):} 10.1109/LRA.2020.2969921}
\thanks{The official IEEE published version of this manuscript can be accessed at: \href{https://ieeexplore.ieee.org/abstract/document/8972387}{https://ieeexplore.ieee.org/abstract/document/8972387} 
}
}

\markboth{Accepted for publication in IEEE Robotics and Automation Letters. Citation information: DOI 10.1109/LRA.2020.2969921.}{Shree \MakeLowercase{\textit{et al.}}: Interactive Natural Language-based Person Search} 


\maketitle

\begin{abstract}
In this work, we consider the problem of searching people in an unconstrained environment, with natural language descriptions. Specifically, we study how to systematically design an algorithm to effectively acquire descriptions from humans. An algorithm is proposed by adapting models, used for visual and language understanding, to search a person of interest (POI) in a principled way, achieving promising results without the need to re-design another complicated model. We then investigate an iterative question-answering (QA) strategy that enable robots to request additional information about the POI's appearance from the user. To this end, we introduce a greedy algorithm to rank questions in terms of their significance, and equip the algorithm with the capability to dynamically adjust the length of human-robot interaction according to model's uncertainty. Our approach is validated not only on benchmark datasets but on a mobile robot, moving in a dynamic and crowded environment.
\end{abstract}


%
\IEEEpeerreviewmaketitle

\section{Introduction}
\label{sec:introduction}

\IEEEPARstart{H}{uman-robot} interaction has attracted increasing attention in recent years. Many robots nowadays are equipped with rich sensors to enable different forms of interactions; for example, a social robot named Zora has cameras, microphones, tactile sensors, position sensors, force sensors and sonar. Among different forms of interactions, using visual and natural language information~\cite{antol2015vqa,wang2015zero,chang2015heterogeneous} is of particular interest and is commonly viewed as the most user-friendly way because of its frequent use in how we humans interact with each other. 

\begin{figure}
\vspace{0.1in}
    \centering
    \includegraphics[trim= 65 80 155 80, clip, width=.42\textwidth]{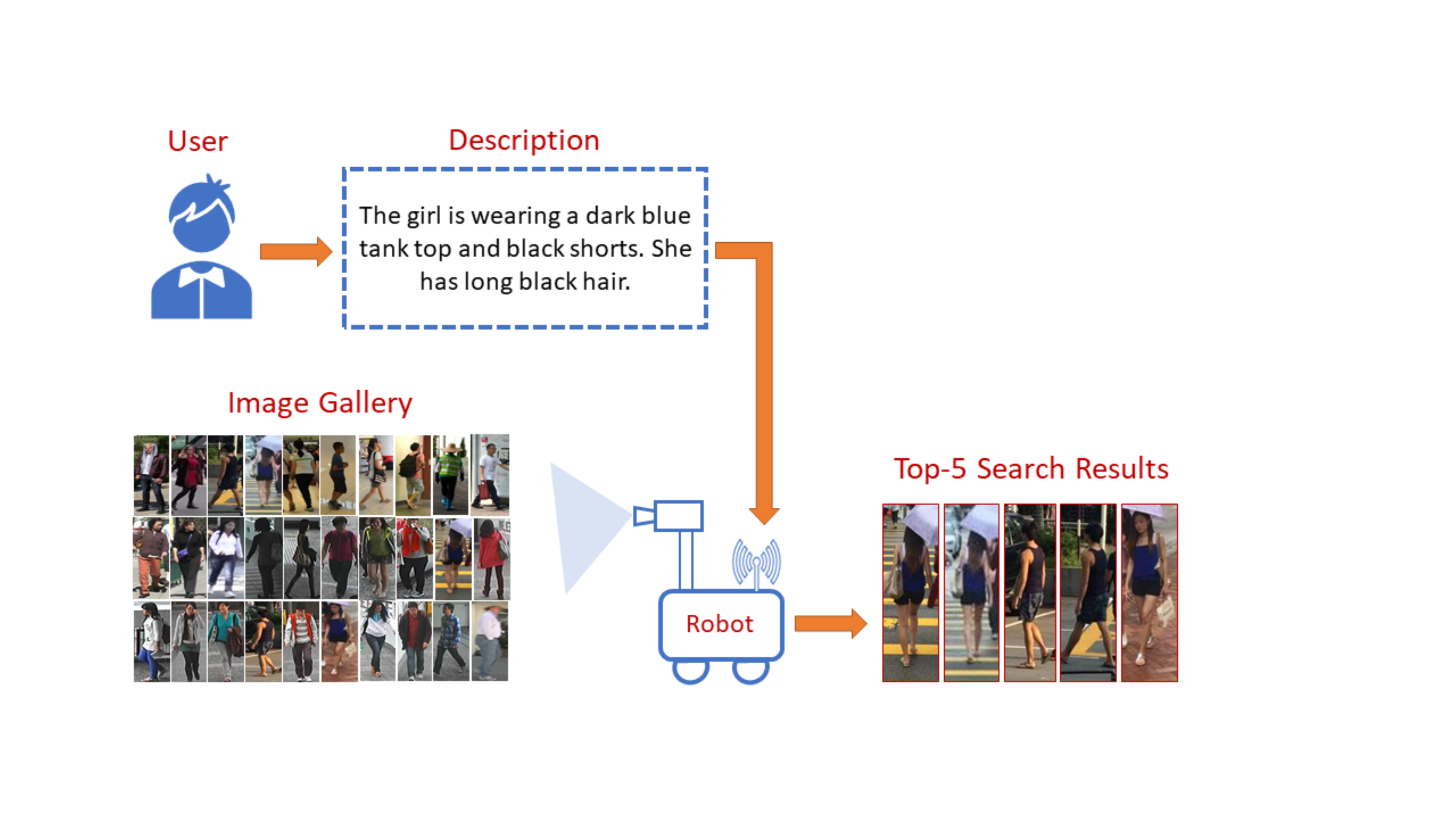}
    \caption{\small Illustration of zero-shot person re-ID. A mobile robot, equipped with a camera, collects images from a crowd to build a gallery. A user provides a language-based description of a target person and the robot returns the top-5 most likely person images.}
    \label{fig:spotlight}
    \vspace{-0.2in}
\end{figure}

In this paper, we study the task of person re-identification (re-ID) which has a great potential to benefit from effective human-robot interaction. In re-ID, a robot is asked to search a target person (in the environment or from a gallery set of images) whose visual appearance may have significant changes due to variations in viewpoints and lightning conditions. Conventional re-ID assumes that a query image of the POI is available and solves the task via similarity matching in the visual domain. There are, however, many practical scenarios such as security, search-and-rescue, in which the assumption is likely infeasible and we have to alternatively rely on verbal descriptions of the POI. Existing literature refers to this task of using descriptions for person search as zero-shot re-ID \cite{roth2014exploration, layne2014attributes,wang2015zero} due to the  missing query images. 
See Figure~\ref{fig:spotlight} for an illustration. 

Zero-shot re-ID, compared to conventional re-ID, has two particular challenges. First, a (query) image itself is worth a thousand words, but it is unlikely to ask a thousand words about the POI from the users. How to acquire the description efficiently is therefore important. Second, in zero-shot re-ID, the robot needs to match the description to images for person search, demanding a 
proper multi-modal similarity measure. For the first challenge, prior work proposed to use a list of visual attributes such as hair-type and clothing to describe the target person's visual appearance~\cite{layne2014attributes, su2016deep, yin2018adversarial} which, however, is time-consuming to annotate and are insufficient to describe a variety of appearance changes. Li \textit{et al.} \cite{li2017person} introduced the CUHK-PEDES dataset for re-ID using natural language descriptions, giving users more flexibility to describe the appearance of a person. Yet, how to obtain informative descriptions to differentiate among different people remains unsolved. For the second challenge, many algorithms have been proposed specifically for the re-ID task~\cite{liao2015person,liao2015efficient}, lacking a connection to the broader literature of visual-language embedding and understanding~\cite{antol2015vqa,chang2015heterogeneous}.


In this paper, we focus on language-based re-ID.
Instead of proposing a new re-ID algorithm from scratch, we argue that language-based re-ID can be viewed as a visual question answering (VQA) task~\cite{antol2015vqa}, in which the input is an image-description pair, and the output is a binary answer (either match or not).
To this end, we propose to modify a leading VQA algorithm named Pythia~\cite{jiang2018pythia} that incorporates LSTM-based sentence embedding and language-guided visual attention and won the VQA Challenge 2018. We show that, with a proper training strategy, our approach can achieve comparable or even better person search accuracy than the state-of-the-art algorithms~\cite{chen2018improving}.

We notice that during language-based retrieval, information provided by the user may not be sufficient to identify the POI if the description is not discriminative enough. 
Consequently, instead of passively giving users the freedom to describe the target person, we investigate a guiding strategy in which the robot actively asks for specific appearance characteristics from the users in a sequential manner. To this end, we define a set of guiding questions that are sufficient to cover a person's appearance, and optimize the order by their significance in reducing the uncertainty in person searching. The resulting robot therefore can interact with users, dynamically asking for additional information if the current description is inadequate to identify the POI.

The main contributions of our work are can be summarized as follows:
\begin{itemize}
    \item First, an algorithm for language-based person search is proposed by properly adapting
    VQA models, achieving competitive accuracy to state-of-the-art methods. 
  
  \item Second, a complementary question-answering dataset (CUHK-QA) was created by designing a set of guiding questions about visual appearances of people. An offline strategy is developed to rank the questions into a sequence to maximize the person search performance.
  Our strategy demonstrates superior performance compared to a randomized baseline strategy in selecting questions.

    \item Third, an information-theoretic scheme is developed to quantify the uncertainty associated with the current person search result, enabling a robot to decide whether to ask for additional information. Our approach therefore allows to trade between re-ID accuracy and the length of human-robot interaction.
    
    \item Finally, we validate our algorithm on a mobile robot, moving in an unconstrained environment. By conducting offline and online studies we show the robustness of our approach in a real-world scenario.

\end{itemize}

\section{Related Work}
\label{sec:relatedWork}
A number of methods have been proposed for zero-shot re-ID that rely upon attributes \cite{layne2014attributes, su2016deep, yin2018adversarial}.
Layne \textit{et al.} \cite{layne2014attributes} propose mid-level attributes to represent discriminating features between person images. Kernelized SVM is used to detect attributes of the gallery images and further matched to the query attributes based on weighted nearest neighbor in the attribute domain. Su \textit{et al.} \cite{su2016deep} develop a semi-supervised deep attribute learning framework where only a small dataset with attribute labels is used to train a deep-CNN model. The network is then fine-tuned on a much larger dataset consisting only of identity labels for people. Yin \textit{et al.} \cite{yin2018adversarial} have developed an adversarial attribute-image re-ID framework to learn semantically discriminative representation in a joint space.

However, Li \textit{et al.} \cite{li2017person} show that attributes have limited expressive ability and motivated the use of natural language descriptions for the search problem. An RNN model, with Gated Neural Attention (GNA) is introduced to evaluate the affinity between sentences and person images. The word-level gating enables the model to assign different weights for different words, in accordance with their significance. To make use of identity annotations in benchmark datasets, Li \textit{et al.} \cite{li2017identity} propose an identity-aware, two-staged CNN-based learning framework for text-to-image retrieval. In the first stage, the network learns to discriminate between different identities by utilizing a cross-modal cross entropy (CMCE) loss. The second stage incorporates a coupled CNN-LSTM network, trained on binary cross-entropy (BCE) loss, outputting matching confidence between the descriptions and images. However, as pointed out in \cite{chen2018improving}, the prior models only account for presence of a word in the descriptions, and its spatial location in the sentence is ignored. Thus, a patch-word matching model is introduced, which captures the affinity between local patches in the image and the words. 

Recently, Antol \textit{et al.} \cite{antol2015vqa} introduced the concept of VQA, where a one-word answer is given for a natural language question about an image. Lu \textit{et al.} \cite{lu2018co} classify the VQA models into two broad categories: free-form region based and detection-based. The former focuses on global visual context in the image, while the latter only processes pre-computed detection regions, thus focusing on foreground objects only. Lu \textit{et al.} \cite{lu2018co} emphasized that a combination of both approaches improve the effectiveness of VQA systems. 
This inspired us to use a hybrid person search module, that looks into both: the detected objects and global context, for finding the text-image affinity score. To the best of our knowledge, none of the prior works study the multi-step information retrieval framework, in the context of person re-ID, thus establishing the novelty of our approach.

\section{Re-Identification with Language Description}
\label{sec:reid_language}
In this section, we present formulation of re-ID problem and describe the architecture of our person search module. 

\subsection{Problem Formulation}
\label{subsec:formulation}

Consider a set $\mathbf{G} = \{g_1,\hdots,g_n\}$, consisting of $n$ images from $K$ distinct people representing the search space for re-ID, often referred to as the gallery. 
Also, consider another set $\mathbf{D} = \{d_1,\hdots,d_m\}$ that represents the set of $m$ query descriptions about people whom we want to search in the gallery.
Assume that the identities corresponding to the images in the gallery and the descriptions in the query set are denoted by, $\mathbf{U} = \{u_1,\hdots,u_n\}$ and $\mathbf{V} = \{v_1,\hdots,v_m\}$, respectively, where $u_i, v_i \in \{1,\hdots,K\}$. 

Given a description $d_i \in \mathbf{D}$, the goal of re-ID is to search for images of the corresponding person within the gallery. To this end, we formulate a two-step strategy where first, a neural network predicts the affinity scores between description $d_i$ and images in the gallery $\mathbf{G}$. Denote the affinity predictor by $\mathbf{f}: \mathbf{D} \times \mathbf{G} \rightarrow \mathbf{T}$, where $\mathbf{T} = [t_{ij}]$ represents the $m \times n$ dimensional score matrix. Second, the most likely image $g_{p^i}$ is selected for association with the description $d_i$, based on the affinity score:

\begin{equation}
    {p}^{i} = \arg\max_{j \in \{1, \hdots, n \}}  \hspace{5pt} \mathbf{f}(d_i, g_j)
\end{equation}

\subsection{Network Architecture}
\label{subsec:network}

\begin{figure}
\vspace{0.1in}
    \centering
    \includegraphics[trim= 0 90 20 150, clip, width=.48\textwidth]{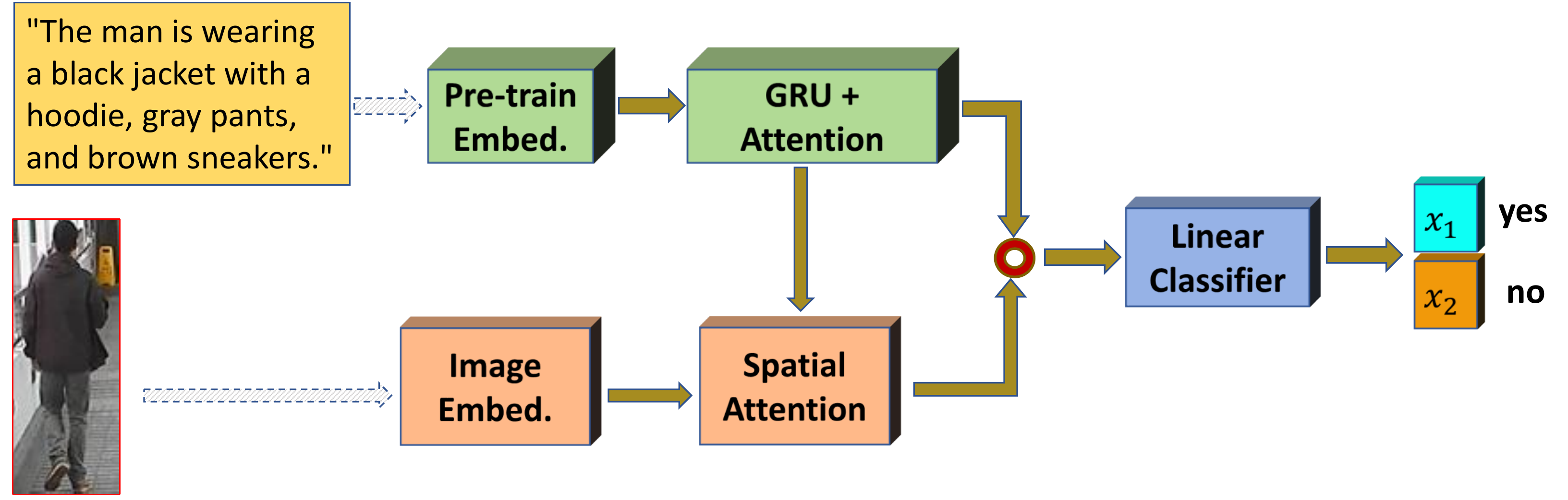}
    \caption{\small Illustration of Pythia-reID person search network architecture. First, image and text features are extracted from the image and description, respectively. Subsequently, it reasons about the similarity of both contents, yielding the likelihoods corresponding to the answers, ``yes'' and ``no''. Light-colored arrows denote pre-trained modules.}
    \label{fig:network}
    \vspace{-0.2in}
\end{figure}

In this work, we leverage an extended Pythia \cite{singh2018pythia}, a state-of-the-art VQA model, for predicting the affinity scores between descriptions and gallery images. Pythia-reID consists of five components, as shown in Figure~\ref{fig:network}:

\renewcommand{\theenumi}{\roman{enumi}}
 \begin{enumerate}
   \item \textbf{Text Embedding}: The words in a description $d_i$ are first embedded with a pre-trained embedding, followed by a gated recurrent unit (GRU) network and an attention module which extracts attentive text features, producing description embedding $\mathbf{f}_{D}(d)$.
   
   \item \textbf{Image Embedding}: A combination of grid and region based features are extracted from an image $g_j$ to encode the image embedding $\mathbf{f}_{I}(g_j)$. As proposed in \cite{jiang2018pythia}, these two types of features capture holistic spatial information about the semantics of the image.
   
   \item \textbf{Spatial Attention}: Based on the image and text features, a top-down attention mechanism outputs a weighted average over spatial features $\mathbf{f}_A(d_i, g_j)$.
   
   \item \textbf{Image-Text Feature Combination}: In order to capture the information common to both, text and image, the attention and text features are combined to obtain the final VQA features $\mathbf{f}_{VQA}(d_i, g_j)$.
   
   \item \textbf{Classifier}: Classifier gives the likelihood of description $d_i$ accurately describing the person in image $g_j$. For re-ID, we modify Pythia to have only two output elements (a) Yes, representing when description corresponds to the person, and (b) No, representing the contrary.
 \end{enumerate}

\subsection{Implementation Details}
\label{subsec:implementation}
In order to extract text features, a pre-trained GloVe embedding \cite{pennington2014glove} is used with a vocabulary of 77k words. A pre-trained ResNet-152 model \cite{he2016deep} obtains 1048-dimensional grid features. The region-based features are obtained from $fc6$ layer of Faster-RCNN model \cite{girshick2015fast}, trained on Visual Genome dataset \cite{krishna2017visual}. A linear layer combines the image and text features, bringing them to the same dimension (5000), followed by an element-wise multiplication and ReLU activation. Finally, in contrast to Pythia which relies on a logistical classifier at its last layer, we use a linear classifier to obtain text-image affinity scores. 

\subsection{Training Scheme}
\label{subsec:implementation}

While training the network, the mean cross-entropy loss with sigmoid activation is minimized:
\begin{equation}
    l(\mathbf{x}, \mathbf{y}) = - \frac{1}{2} \sum_{k=1}^{2} y_k \log{(\sigma{(\mathbf{x})_k})} + (1-y_k) \log{ (1 - \sigma (\mathbf{x})_k) }
\end{equation}
where $\sigma(\mathbf{x})_k$ denotes the predicted likelihood for $k^{th}$ output after applying a \textit{softmax} function and $y_k$ denotes the corresponding ground truth label (`Yes' $\mathbf{y}$ = [1,0] and ``No'' $\mathbf{y}$ = [0,1]). 
Positive and negative samples of sentence-image pairs are used while training, where a positive sample represents sentence-image pairs corresponding to each other and a negative sample represents the contrary.
Each pair is randomly sampled from the dataset with a ratio of 2 positive : 3 negative. 

\subsection{Dataset and Evaluation Metrics}
\label{subsec:data_metric}
The performance of our Pythia-reID module is evaluated on CUHK-PEDES, a language-based person search dataset. It consists of 40,206 images of 13,003 people and two independent language descriptions about each image. Similar to \cite{li2017person}, the training set consists of 33,987 images of 11,003 people with 67,974 sentence descriptions. The validation and test sets are comprised of 3,078 and 3,074 images, respectively, both with 1,000 people in it.

As standard in re-ID, the top-$k$ accuracy is used as performance metric. For a given sentence description, the affinity score is calculated for the entire image gallery and the images are ranked in the order of decreasing affinity. A successful search is accomplished if the person of interest is among the top-$k$ images from the sorted gallery. 

\begin{figure*}
\centering
  \begin{subfigure}[b]{0.47\textwidth}
    \includegraphics[trim= 160 200 200 50, clip,width=0.8\textwidth]{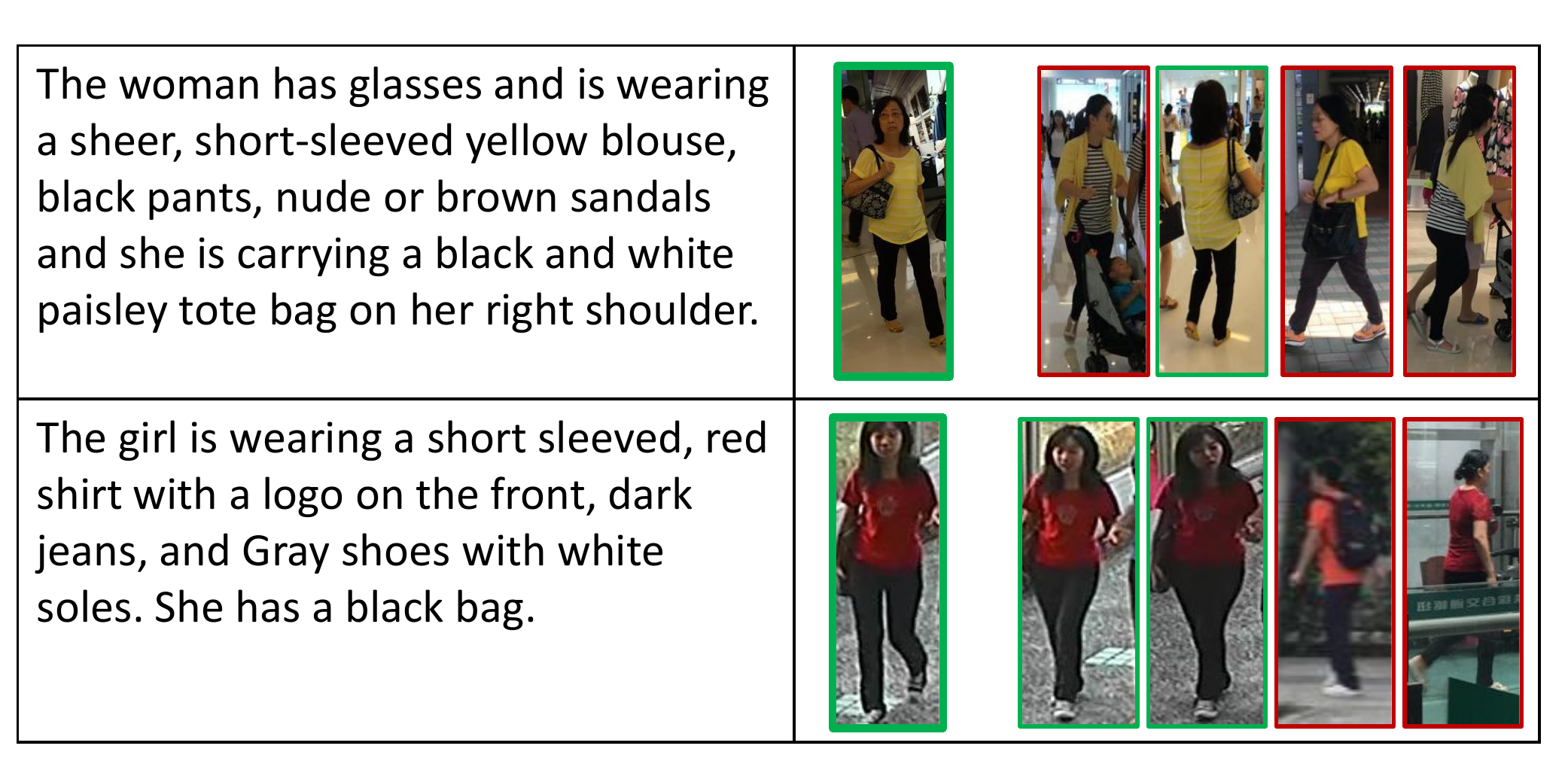}
    \caption{\small Retrieval results when top-$1$ prediction matches ID of POI.}
    \label{fig:goodResults}
  \end{subfigure}
  \ \ \ \  
  \begin{subfigure}[b]{0.48\textwidth}
    \includegraphics[trim= 160 205 200 50, clip,width=0.8\textwidth]{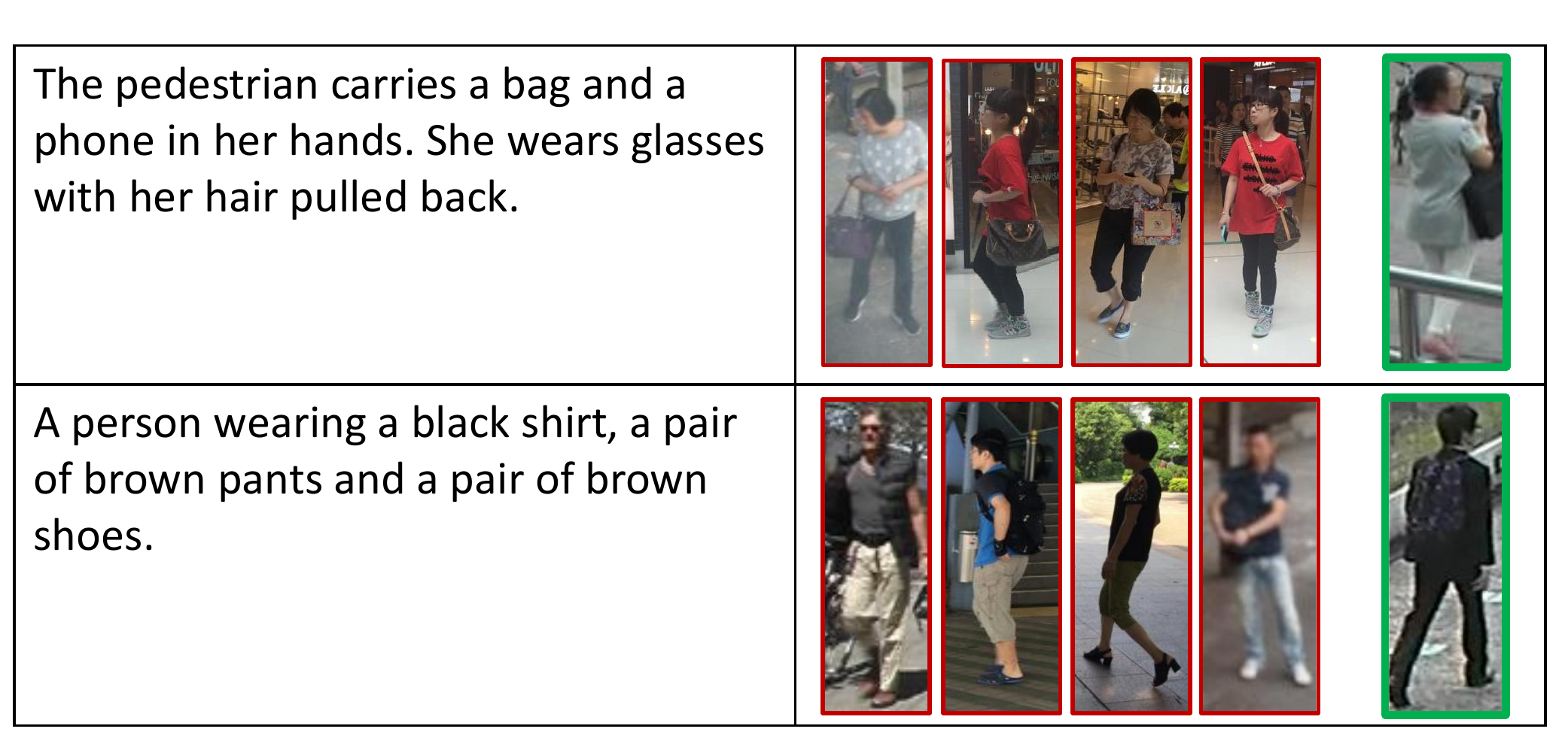}
    \caption{\small Retrieval results when top-$1000$ predictions do not include POI.}
    \label{fig:badResults}
  \end{subfigure}
  \caption{\small Qualitative person-search results on CUHK-PEDES. Green box implies image belongs to POI, while red box implies the contrary.}
  \label{fig:qualResults}
  \vspace{-0.2in}
\end{figure*}

\subsection{Results}
\label{subsec:pythia_results}

In Table~\ref{tbl:languageReid}, the proposed Pythia-reID model is compared with other state-of-the-art language-based person search frameworks.
Pythia-reID achieves the best top-$10$ accuracy and the second best top-$1$ accuracy, affirming the relevance of VQA models for language-based re-ID. Figure~\ref{fig:qualResults} presents some qualitative retrieval results of Pythia-reID to provide deeper insight. For the successful cases, shown in figure~\ref{fig:goodResults}, the best retrieval images contain significant overlap in terms of appearance attributes described in the description. 

\begin{table}
\vspace{0.1in}
\centering
\caption{ \small Comparison of the top-\textit{k} accuracy of language-based person search for different models.}
\begin{tabular}{|p{3.5cm}|p{1.2cm}|p{1.2cm}|}
\hline
\textbf{Models} & \textbf{Top-1} & \textbf{Top-10}\\ \hline
 QAWord 2015 \cite{antol2015vqa} & 11.62 & 42.42\\
 \hline
 NeuralTalk 2015 \cite{vinyals2015show} & 13.66 & 41.72\\
 \hline
 GNA-RNN 2017 \cite{li2017person} &  19.05 & 53.64\\
 \hline
 Two-stage 2017 \cite{li2017identity} &  25.94 & 60.48\\
 \hline
 PWM+ATH 2018 \cite{chen2018improving}& \textbf{27.14} & 61.02\\
 \hline
 Pythia-reID (Ours) &  26.79 & \textbf{63.78}\\
 \hline 
\end{tabular}
\label{tbl:languageReid}
\vspace{-0.1in}
\end{table}

We also studied a few failure cases where the POI is not present in the top-$1000$ retrieval results, and found that lack of discriminative information is a major reason for such behavior. For example, predictions in Figure~\ref{fig:badResults}(top) include images with a woman who is carrying bag and wearing glasses, however, the dress type and color, which is one of the decisive factors for the search module, is absent in the description. This example aptly highlights the opportunity for starting a conversation between the robot and the user, with the goal of improving the retrieval results and motivates the next section where a multi-step information retrieval process for language-based re-ID is proposed.

\section{Supervised Information Retrieval for Person Search}
\label{sec:supervised_search}

Until this point, the retrieval method relies only upon the initial description provided by the user. In this section, we propose a novel, sequential QA scheme where the user is requested to respond to a sequence of questions describing the appearance of the person of interest, in order to improve search performance. This methodology of information retrieval enables more distinct descriptions to be acquired about person's appearance. It also serves as the foundation for models that can choose to ask additional questions, if the current description is insufficient.

\subsection{A Greedy Strategy for Information Retrieval}
\label{subsec:greedy_algo}
Not all the query questions are equally valuable for the person search problem. Certain aspects of a person's appearance can be more distinctive than others. For example, ``dark shoes'', may not be as useful as ``bright yellow top'', since dark shoes are usually more common. Thus, an intelligent QA system should schedule the questions in a sequence that maximizes the person-retrieval performance.

Consider a set of $n_Q$ questions $\mathbf{Q} = \{q_k\} \ \forall k \in \{1,\hdots,n_Q\}$, about the appearance. For a given image $g_i$ from the gallery $\mathbf{G}$, $n_Q$ descriptions correspond to the questions in $\mathbf{Q}$, denoted by $\mathbf{d}_i = \{d_i^1, d_i^2, \hdots, d_i^{n_Q}\}$; also, $\mathbf{D} = \{\mathbf{d}_1, \hdots, \mathbf{d}_n\}$. Assume that $\mathbf{S}$ is a list, representing the order of descriptions. Thus, $\mathbf{S} = [s_k]$, where $s_k \in \{1, \hdots, n_Q\}$, such that $s_k = s_l \iff k=l$. Define a metric, \textit{rank}, that represents the minimum index of a person image corresponding to the POI in the gallery, which is sorted according to decreasing text-image affinity scores. 
The \textit{rank} represents the number of incorrect retrieval results that our model outputs before a correct image corresponding to the POI; thus, having a lower \textit{mean rank} ($M$) implies better retrieval performance. 
The goal is to prioritize the questions in decreasing order of significance.
To pose the task as a maximization problem, another metric: $R(\mathbf{S},\mathbf{D},\mathbf{G}) = n - M(\mathbf{S},\mathbf{D},\mathbf{G})$ is defined, where $n$ is the size of the gallery. The maximization goal is to find $\mathbf{S}^{*}$, such that:
\begin{equation}
    \vspace{-0.05in}
    \mathbf{S}^{*} = \arg\max_{\mathbf{S}} \hspace{5pt} R(\mathbf{S},\mathbf{D},\mathbf{G})
    \label{eq:optimize_greedy}
\end{equation}

Unfortunately, the search space of Equation~\ref{eq:optimize_greedy} is huge for large $n_Q$.
Noting that heuristic solutions could work well for iterative QA, we propose a greedy algorithm (Algorithm 1) that iteratively chooses the question $q_{k}$ in order to maximize the performance $R$ at the current information retrieval step. While, greedy algorithms can result in an arbitrarily poor solution to optimization problems, Nemhauser \textit{et al.} \cite{nemhauser1981maximizing} prove that the solution obtained from the greedy algorithm is a good approximation for the optimal solution if the objective function ($R$) is a submodular. Since $R$ is a function of output of a deep neural network, proving its submodularity property is a challenge in general. However, for an \textit{ideal} person search module\footnote[3]{Given description set $\mathbf{d}_i = \{d_{i}^{1}, \hdots, d_{i}^{n_Q}\}$ and gallery $\mathbf{G}$, we define an \textit{ideal} person search module as the one that accurately outputs an image set $\mathbf{g}_{i} \subseteq \mathbf{G}$, satisfying the appearance descriptions of $\mathbf{d}_i$, irrespective of the sequence $\mathbf{S}$ in which descriptions are arranged. For example, if $\mathbf{d}_i = $\{``person has red shirt'', ``He wears jeans''\}, the output contains all images of men wearing jeans and red shirt. Since all output images $\mathbf{g}_i$ are equally likely, thus the \textit{rank} for an \textit{ideal} person search module can be reasonably defined as $\frac{|\mathbf{g}_i| + 1}{2}$.}, we prove that $R$ indeed satisfies the submodularity criteria (Appendix\ref{sec:appendix}). 
Thus, while not provably optimal, it is reasonable to assume that for a superior person search network a solution obtained from Algorithm~\ref{alg:greedy_algorithm} is a good approximation of the optimal solution.

\begin{algorithm}[H]
 \textbf{Data: }{Image gallery $\mathbf{G}$ and description set $\mathbf{D}$, consisting of $n_Q$ sentences corresponding to each question in $\mathbf{Q}$.}\\
 \textbf{Result: }{Sequence of descriptions $\mathbf{S}$.}\\
 $\mathbf{S} = [ \ ] $\\
 \textbf{While }({$|\mathbf{S}| < n_Q$}):\\
 \ \ {\textbf{comment:} $\oplus$ denotes list merging operation\;\\
  \ \ $i^{*} \gets \arg \min_{i \in \{1, \hdots, n_Q \} \backslash \mathbf{S}} \hspace{5pt} M(\mathbf{S} \oplus [i],\mathbf{D},\mathbf{G})$\;\\
  \ \ $\mathbf{S} \gets \mathbf{S} \oplus [i^{*}]$
 }\\
 \textbf{end}\\
 \textbf{return }{$\mathbf{S}$}
 \caption{Greedy Algorithm for Iterative Info-Retrieval}
 \label{alg:greedy_algorithm}
\end{algorithm}

\subsection{Dataset for Supervised Person Search}
\label{subsec:participant_data}

Since there is no existing dataset for natural language-based person search using iterative questioning, we built our own benchmark. To this end, we opt for five query questions about an image, each asking about certain appearance aspect of a person. Table~\ref{tbl:questions} shows the questions, which are motivated from the recent work in attribute-based re-ID \cite{shree2019empirical}. We randomly selected 400 images of 360 people from the test set of CUHK-PEDES dataset, and conducted a survey to label the images with answers to the query questions (named CUHK-QA). To encompass diversity in the language descriptions, we recruited 20 participants for the survey, each labelling 20 images. All the participants were graduate students, enrolled at Cornell University.

\begin{figure}
    \centering
    \includegraphics[trim= 10 370 85 0, clip, width=.49\textwidth]{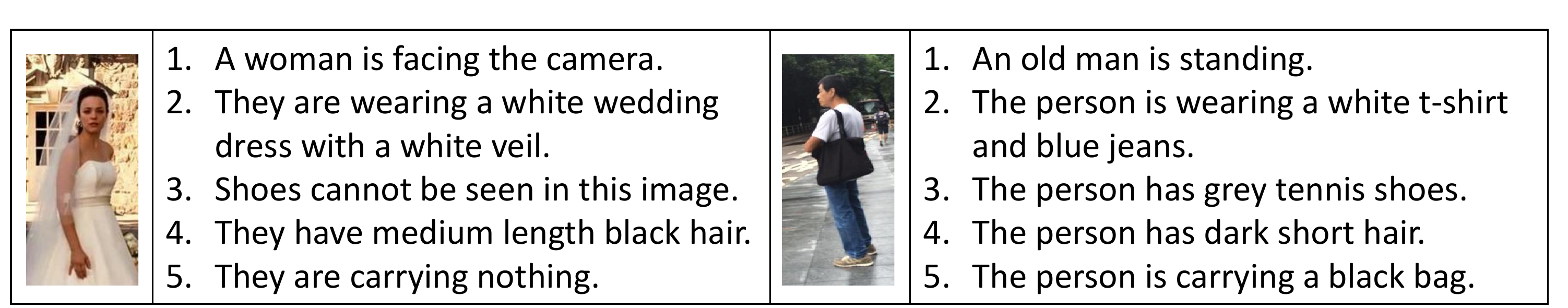}
    \caption{\small A few samples from our QA dataset (CUHK-QA), describing persons' appearance based on the questions in Table~\ref{tbl:questions}.}
    \label{fig:participantDataset}
    \vspace{-0.2in}
\end{figure}

The dataset consists of 2,000 high quality sentence labels, describing the appearance of a person. 
The average length of the combined description per image is 39.15 words, which is significantly higher than CUHK-PEDES dataset, where the average sentence length is 23.5. The labelled dataset has been open sourced\footnote[4]{The labelled data collected from the study can be viewed at \href{https://github.com/vikshree/QA_PersonSearchLanguageData}{\texttt{https://github.com/vikshree/QA\textunderscore PersonSearchLanguageData}}}  to promote research in language-based re-ID. Figure~\ref{fig:participantDataset} shows a few samples from the collected data.


\subsection{Evaluation}
\label{subsec:greedy_eval}
To validate our strategy of using greedy algorithm for optimizing question sequences, we divided the CUHK-QA dataset into two halves: train and test, with 200 images in each. In addition, we ensured that images in each are labelled by different participants. 
To evaluate the \textit{mean rank} $M$ in Algorithm~\ref{alg:greedy_algorithm}, the fused descriptions in $\mathbf{D}$, as per the current sequence $\mathbf{S}$ are fed, to the person search module (Pythia-reID). A simple sequential concatenation operation is performed for fusing descriptions. 

\begin{table}
\vspace{0.1in}
\centering
\caption{\small Questions used for describing a person's appearance.}
\begin{tabular}{|lp{7.3cm}|l|l|}
\hline
 \textbf{$\#$} & \textbf{Questions}\\ \hline
 1. & Describe gender of the person, age group and any action they are involved in. \\ \hline
 2. & Describe appearance of dress that the person is wearing. \\ \hline
 3. & Describe footwear of the person. \\ \hline
 4. & Describe appearance of person’s hair including color and length. \\ \hline
 5. & Describe other accessories that person might be wearing or carrying or holding. \\ \hline
\end{tabular}
\label{tbl:questions}
\vspace{-0.1in}
\end{table}

\begin{figure}[h]
\vspace{-0.1in}
\centering
  \begin{subfigure}[b]{0.21\textwidth}
    \includegraphics[trim= 0 10 16 25, clip,width=\textwidth]{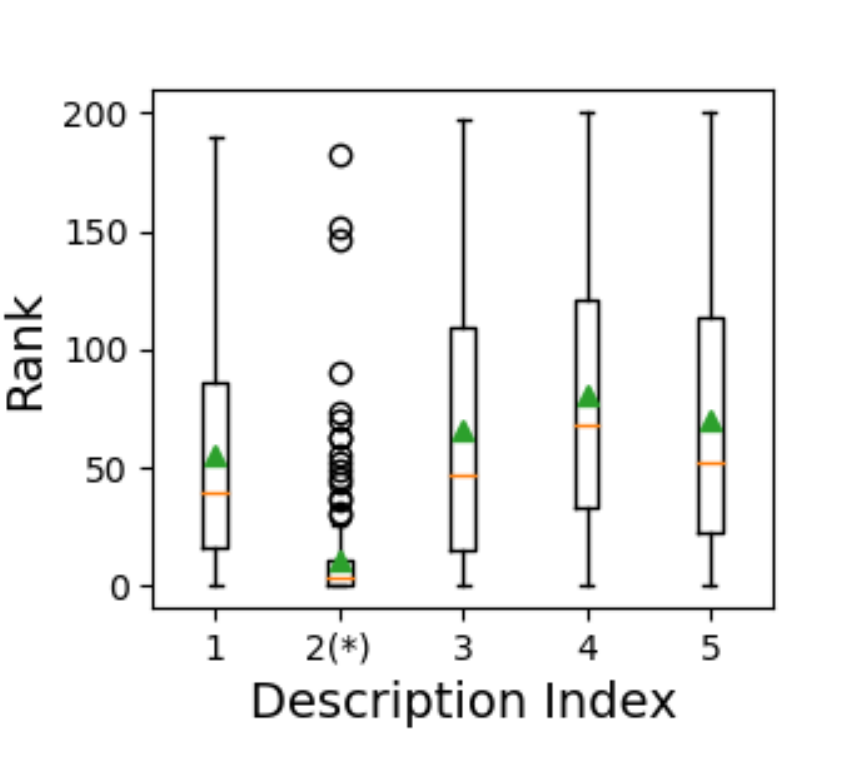}
    \caption{\small Iteration-1}
  \end{subfigure}
  \ \ \
  \begin{subfigure}[b]{0.21\textwidth}
    \includegraphics[trim= 0 10 16 25, clip,width=\textwidth]{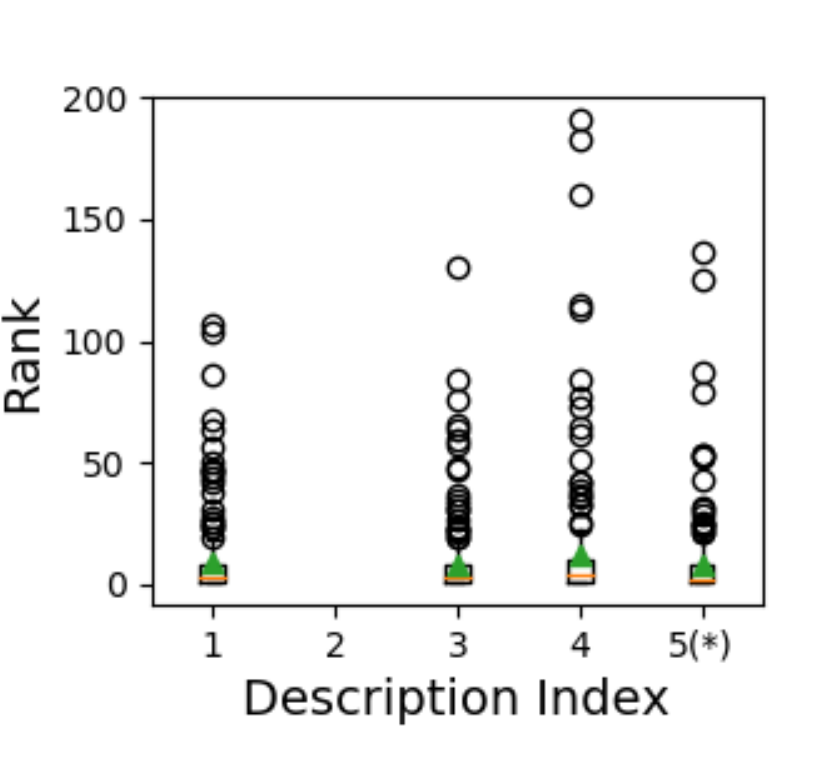}
    \caption{\small Iteration-2}
  \end{subfigure}
  \caption{\small Distribution of \textit{rank} performance at each step of Greedy Algorithm~\ref{alg:greedy_algorithm}. The description corresponding to minimum $M$, denoted by (*), is chosen at the end of every iteration.}
  \label{fig:greedy_Steps}
  \vspace{-0.1in}
\end{figure}

For optimizing the sequence of questions, we start with Pythia-reID model, trained on CUHK-PEDES training set, following the steps of the Algorithm~\ref{alg:greedy_algorithm} to iteratively select the questions that maximize the performance in Equation~\ref{eq:optimize_greedy}. Figure~\ref{fig:greedy_Steps} depicts a few steps of Algorithm~\ref{alg:greedy_algorithm} and the obtained sequence is: 
\begin{equation}
    \mathbf{S}^{'} = [2, 5, 1, 3, 4]
    \label{eq:greedy_sol}
\end{equation}

\begin{figure}
\centering
\vspace{0.1in}
  \begin{subfigure}[b]{0.23\textwidth}
    \includegraphics[trim= 10 20 20 30, clip,width=\textwidth]{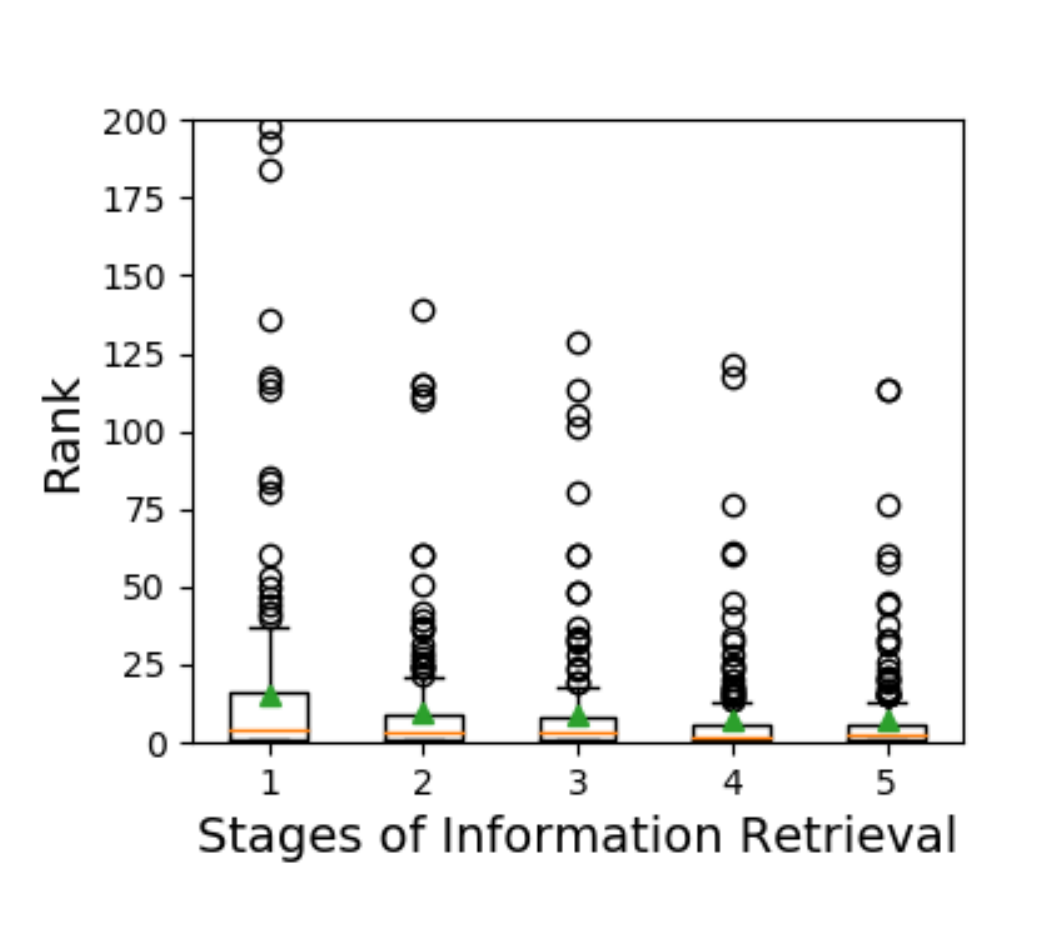}
    \caption{\small Optimized question order $\mathbf{S}^{'}$}
  \end{subfigure}
  \ \ \
  \begin{subfigure}[b]{0.23\textwidth}
    \includegraphics[trim= 10 20 20 30, clip,width=\textwidth]{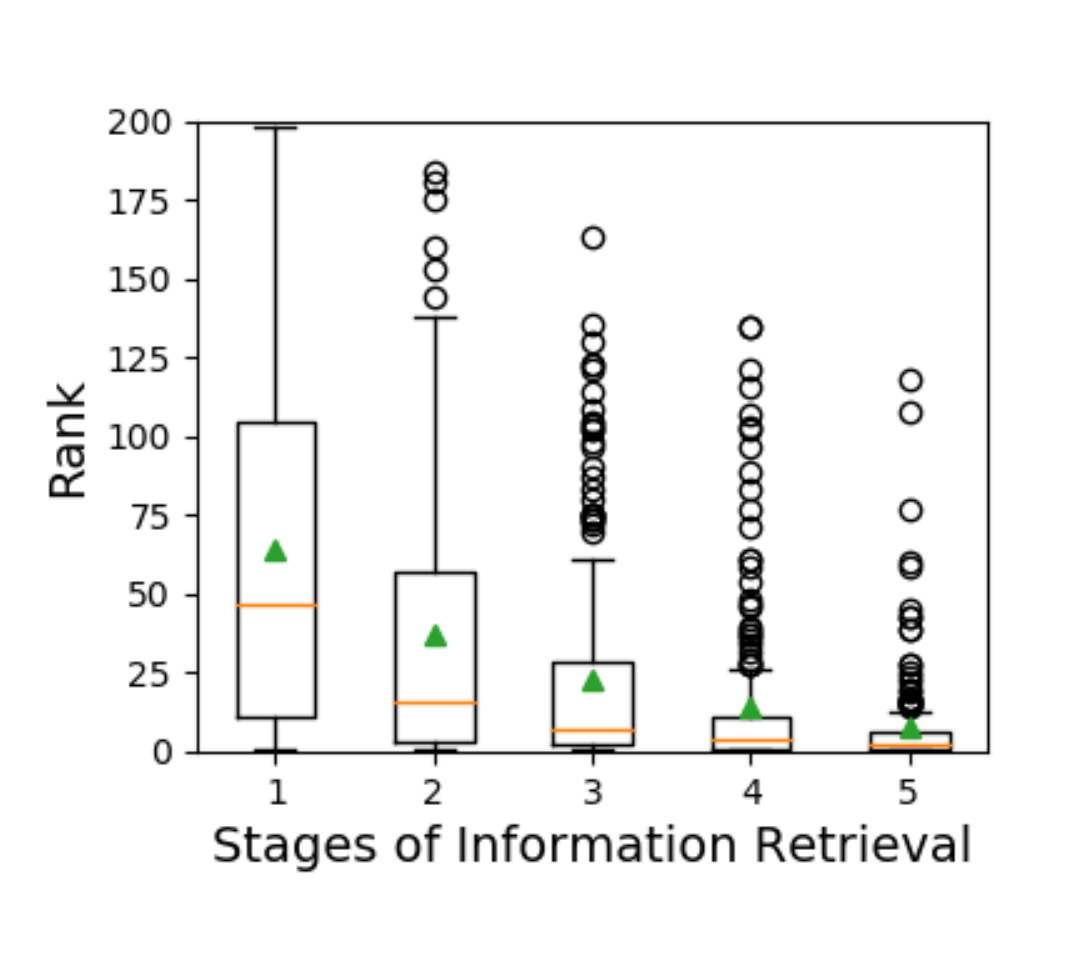}
    \caption{\small Random question order}
  \end{subfigure}
  \caption{\small Comparison of \textit{rank} performance if questions as $\mathbf{S}^{'}$ versus asking them in a random sequence. Gallery has 200 images.}
  \label{fig:compare_greedy_random}
  \vspace{-0.2in}
\end{figure}

We test the learnt strategy on CUHK-QA test set and compare it against a randomized strategy, where a random sequence is picked for each image; the results are shown in Figure~\ref{fig:compare_greedy_random}. We observe that although the final \textit{mean rank} achieved in both cases is about $7$, yet, the greedy algorithm converges much faster to this value. A key conclusion is that with a fixed allowance on the number of questions, we should ask them in the order $\mathbf{S}^{'}$ to achieve superior performance.

\begin{figure*}
\vspace{0.1in}
    \centering
    \includegraphics[trim= 20 340 70 5, clip, width=0.95\textwidth]{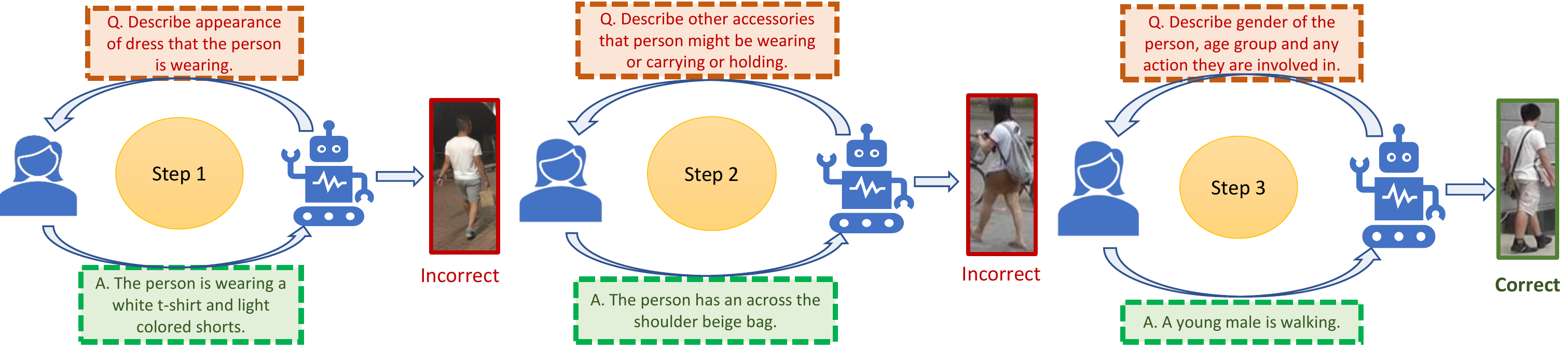}
    \caption{\small A sample result for supervised information retrieval based on Algorithm~\ref{alg:greedy_algorithm}. At each step, the user is asked a question relating to the appearance of the POI, and the corresponding top-\textit{1} search result is shown.}
    \label{fig:qaResult}
    \vspace{-0.2in}
\end{figure*}

Figure~\ref{fig:qaResult} shows a typical example of how iterative QA could help in improving performance. In step(1), based on the description, the search module selects a person wearing similar outfits to the POI. In step(2), our algorithm asks about accessories; however, the gender of the retrieved person is different than the POI. At last, having sufficient information, the search module outputs a correct image.

\section{Uncertainty driven Information Retrieval}
\label{sec:uncertaintyInfo}

The complexity of searching for someone in a gallery varies from person to person. For example, a circus clown, wearing a colourful gown is much easier to identify than a person wearing a blue suit, in a dataset collected from an airport. Thus, the number of questions required to search different persons should be different, depending upon the distinctive attributes of the person. In this section, we propose to leverage the uncertainty in the prediction and the information content of the description to decide whether additional information is required for identifying the POI or not.

\subsection{Quantifying Uncertainty in Predictions}
\label{subsec: quantifyUncertainty}

In information theory, entropy is often used to characterize uncertainty. 
In the context of person search, using text description, text-image affinity scores that are very close to each other implies that the network is highly uncertain about its prediction, and vice-versa. As an example, consider a gallery of images where most people are wearing black pants. A text description: ``The person is wearing black colored pants.", would result in very similar affinity scores for the entire gallery; thus, little information is gained and practically, it is still challenging to identify the correct POI. In such a situation, further information should be requested from the user. 
From Section~\ref{sec:supervised_search}, we already know the sequence of questions to ask to the user that would optimize the information gain.
Here, we propose to treat entropy of the affinity score distribution as a metric for uncertainty in the predictions. Furthermore, we hypothesize a threshold based approach, where the robot continues to ask questions that yield more information until a pre-specified entropy level is achieved; this level is referred to as the \textit{budget of uncertainty}.

Given a set of descriptions $\mathbf{d}_{i}$ about an image $g_i$, denote the corresponding affinity-score for the entire gallery as $\mathbf{A}_{i} = [a_{i,j}]$, such that $a_{i,j} \in [0,1] \ \forall j \in \{1,\dots, n\}$.
By normalizing the scores, a probability distribution over the gallery is realized, denoted by $\mathbf{\hat{A}}_{i} = [\hat{a}_{i,j}]$, such that $\hat{a}_{i,j} = \frac{a_{i,j}}{\sum_{j=1}^{n} a_{i,j}}$. The entropy of this distribution is:
\begin{equation}
    E_{i} = - \sum_{j=1}^{n} \hat{a}_{i,j} \log( \hat{a}_{i,j} ) 
    \label{eq:entropy}
\end{equation}

\subsection{Evaluation}
\label{subsec: budgetUncertainty}

We evaluate our uncertainty based information retrieval approach on our CUHK-QA dataset, defined in Section~\ref{subsec:participant_data}. 
By default, the QA starts with the first question in the optimized sequence in Equation~\ref{eq:greedy_sol}, and the decision to ask the next question is made on the basis of current uncertainty in the predictions.
Since the dataset consists of five descriptions for each image, a five-step information retrieval process can be simulated. There are 200 images in the test set; thus our system can make a maximum of $4\times200 = 800$ additional queries to the user. 

\begin{figure}[h]
\centering
  \begin{subfigure}[b]{0.23\textwidth}
    \includegraphics[trim= 5 20 20 20, clip,width=0.9\textwidth]{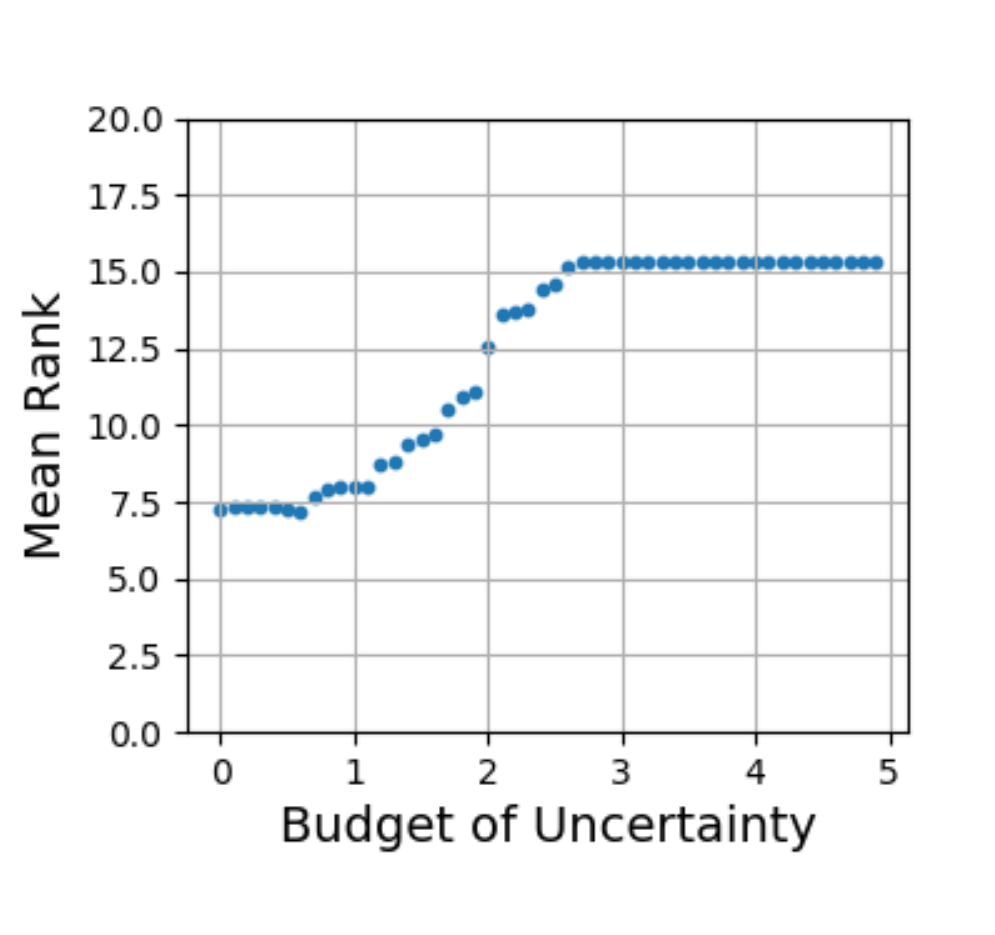}
    \caption{\small \textit{Mean rank} ($M$)}
    \label{fig:meanRank_vs_budget_qa}
  \end{subfigure}
  \ \ \
  \begin{subfigure}[b]{0.23\textwidth}
    \includegraphics[trim= 15 30 20 20, clip,width=0.9\textwidth]{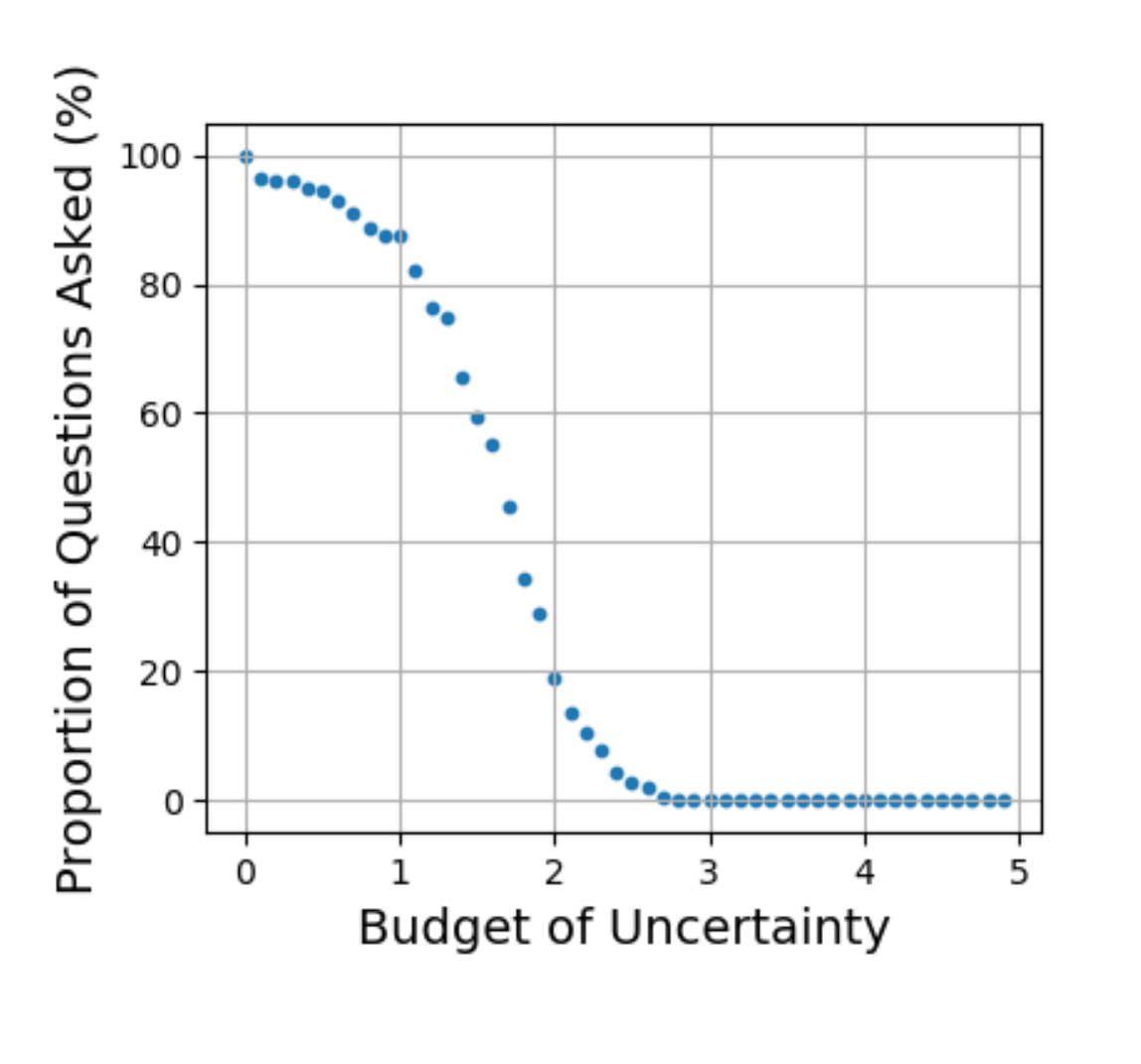}
    \caption{\small Number of queries}
    \label{fig:numQs_budget_qa}
  \end{subfigure}

  \caption{\small Five-step information retrieval results on CUHK-QA dataset for different \textit{budgets of uncertainty}.}
  \label{fig:budget_uncertainty_with_qa}
  \vspace{-0.2in}
\end{figure}

We utilize our pre-trained model of Pythia-reID from Section~\ref{sec:reid_language} for evaluation; results are shown in Figure~\ref{fig:budget_uncertainty_with_qa}. In Figure~\ref{fig:numQs_budget_qa}, we observe that smaller budgets of uncertainty leads to a high query rate for the user, and vice-versa.
Figure~\ref{fig:meanRank_vs_budget_qa} shows that a more stringent uncertainty budget leads to better \textit{mean rank}, because more information is received from the user. In other words, being overly conservative (small uncertainty budget) improves the \textit{mean rank} by about $50\%$, but at the cost of large number of queries. 

\subsection{Discussion}
\label{subsec: uncertainty_discussion}

Based on the results in Figure~\ref{fig:budget_uncertainty_with_qa}, we conclude that our uncertainty based approach of requesting additional information allows a trade between the \textit{mean rank} retrieval performance $M$ and number of queries asked of the user. 
A smaller \textit{budget of uncertainty} can be acquired, leading to lower \textit{mean rank}, but at the cost of additional questions for the user; thus, the ultimate threshold is application dependent. Nonetheless, our framework rescues the user from answering a potentially exhaustive list of questions about the POI's appearance. 

\section{Robotic Experiment}
\begin{figure}[h]
\centering
  \begin{subfigure}[b]{0.13\textwidth}
    \includegraphics[trim= 30 30 860 360, clip,width=0.8\textwidth, height=1.7cm]{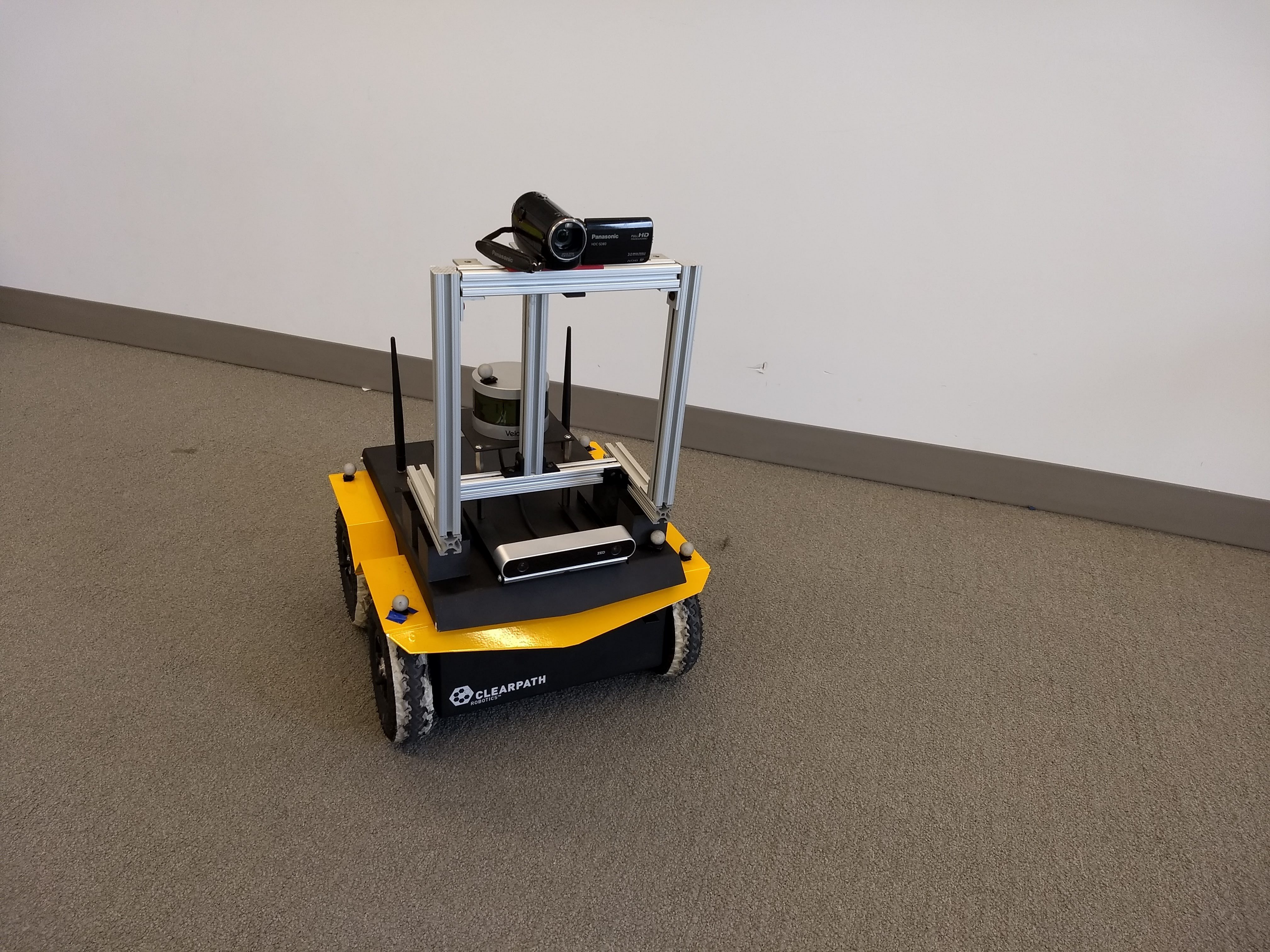}
    \caption{\small }
    \label{fig:exp_robot}
  \end{subfigure}
  \ \ \ 
  \begin{subfigure}[b]{0.21\textwidth}
    \includegraphics[trim= 0 0 180 0, clip,width=0.8\textwidth, height=1.7cm]{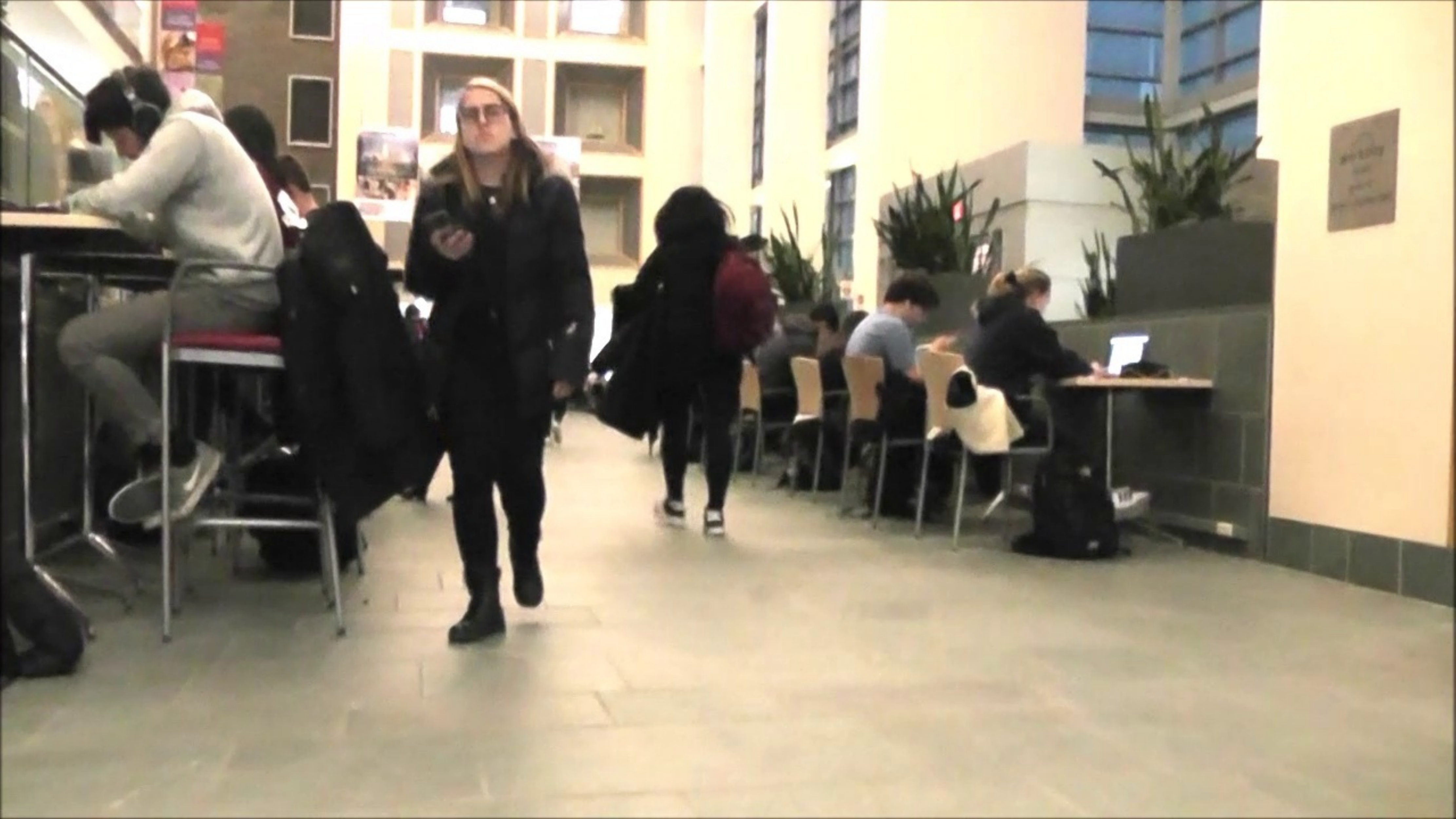}
    \caption{\small }
    \label{fig:exp_snippet}
  \end{subfigure}

  \caption{\small (a) Jackal robot equipped with camera that is used for the experiment. (b) An image-snippet from the video recorded by the robot during experiment in the Duffield Hall at Cornell University.}
  \label{fig:experiment_setup}
  \vspace{-0.2in}
\end{figure}

To validate the practical performance of our QA strategy, we conducted studies with a robot and camera sensor, collecting data in an unstructured environment. Figure~\ref{fig:exp_robot} shows the robot-platform and Figure~\ref{fig:exp_snippet} shows an image-snippet taken from the scene. 
 The venue represents a densely crowded, dynamic environment, making it appropriate for investigating the robustness of our approach.
We conduct both offline (A) and online (B) analysis.

\subsection{Offline Experiment}
In the offline setting, the robot first conducts an exploratory survey of the field and collects video data for 2 minutes, which is post-processed afterwards for finding the POI in the scene. The images are extracted at a rate of 10 FPS and Mask-RCNN network \cite{he2017mask} is used for detecting humans in each frame, with a minimum confidence threshold of 0.98. Subsequently, a gallery of 4,340 bounding boxes is obtained with multiple images corresponding to more than 40 different people in the scene. From the scene, five different people are chosen as POI, exhibiting different configurations including standing, sitting, walking towards or away from the robot. The images corresponding to a POI are hand-labelled to create the ground-truth. Human trials were conducted with five participants; each was asked to answer the questions in Table~\ref{tbl:questions} for two POIs in the scene. Thus, two independent descriptions for each POI were received.

\begin{figure}[h]
\vspace{-0.1in}
\centering
  \begin{subfigure}[b]{0.23\textwidth}
    \includegraphics[trim= 5 20 20 30, clip,width=0.95\textwidth]{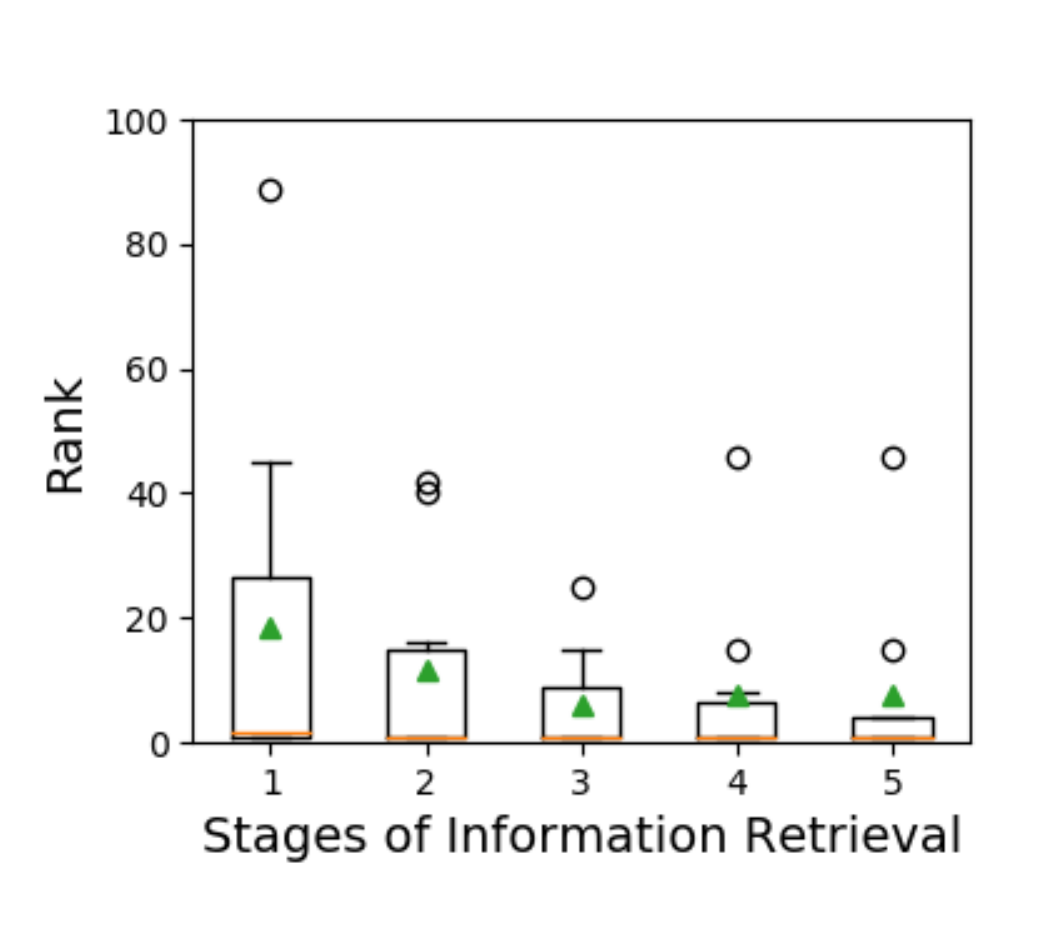}
    \caption{\small Optimized question order $\mathbf{S}^{'}$.}
    \label{fig:offline_results_mean_rank}
  \end{subfigure}
  \ \ \ 
  \begin{subfigure}[b]{0.23\textwidth}
    \includegraphics[trim= 5 20 20 30, clip,width=0.95\textwidth]{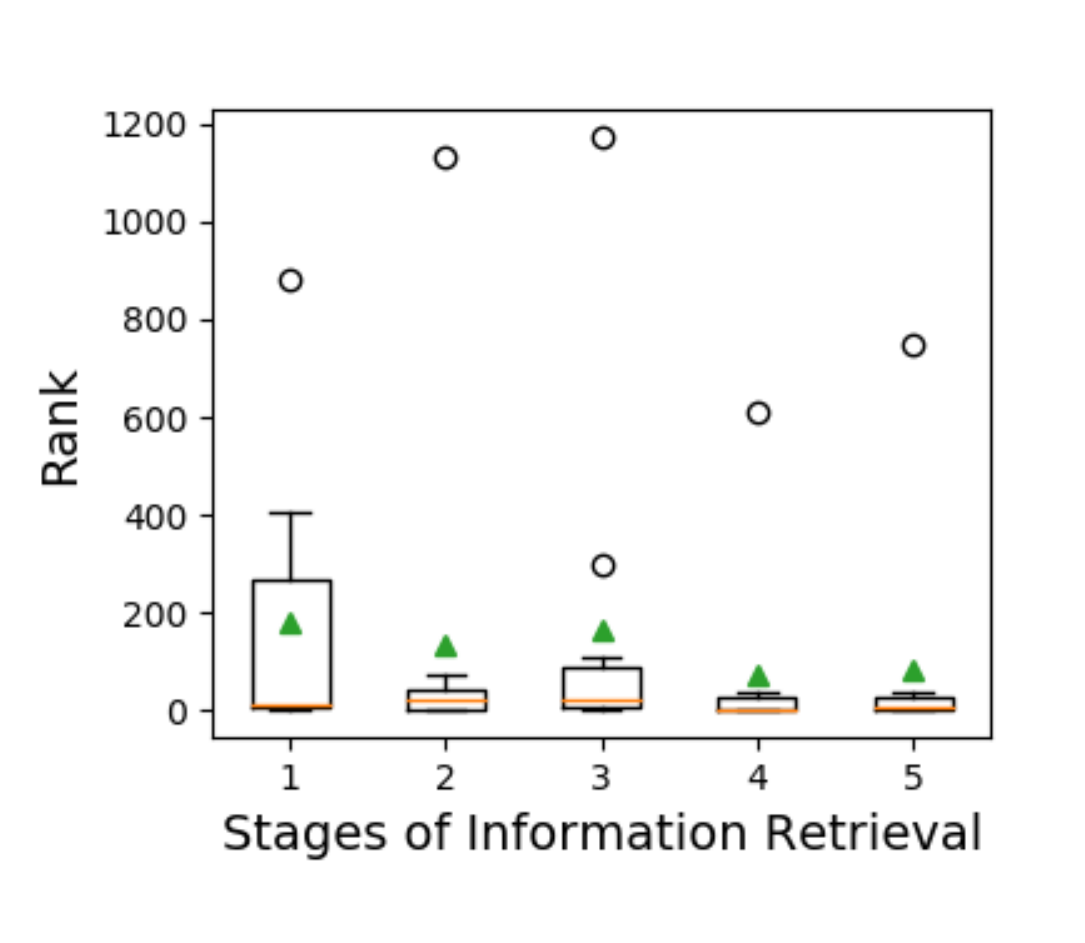}
    \caption{\small Random question order.}
    \label{fig:offline_results_mean_rank_random}
  \end{subfigure}

  \caption{\small Comparison of \textit{rank} performance if questions are arranged as $\mathbf{S}^{'}$ versus asking them in random order. Gallery has 4,340 images. \textbf{Note:} The y-axis has different scales in (a) and (b).}
  \label{fig:offline_results_qa}
  \vspace{-0.1in}
\end{figure}

We first study the \textit{rank} performance achieved by asking the questions in the optimized sequence as obtained from Equation~\ref{eq:greedy_sol}; results are shown in Figure~\ref{fig:offline_results_mean_rank}. We observe that asking more questions leads to consistent reduction in the interquartile range of \textit{rank} performance. For a gallery size of 4,340, the maximum \textit{rank} obtained after asking all the five questions is within $1\%$ of the gallery size, while the \textit{mean rank} $M$ is within $0.25\%$ of the gallery size. Also, after the third question, $M$ saturates at a fixed value, indicating that later questions about the appearance like hair-color and footwear, do not help further disambiguate the POI among the crowd. 
Figure~\ref{fig:offline_results_mean_rank_random} depicts that asking the questions in a randomized order leads to high variance in the \textit{rank} and inferior performance as compared to Figure~\ref{fig:offline_results_mean_rank}, noting that scales of y-axis are different.

\begin{figure}[h]
\centering
\vspace{-0.1in}
  \begin{subfigure}[b]{0.23\textwidth}
    \includegraphics[trim= 5 10 25 0, clip,width=0.9\textwidth]{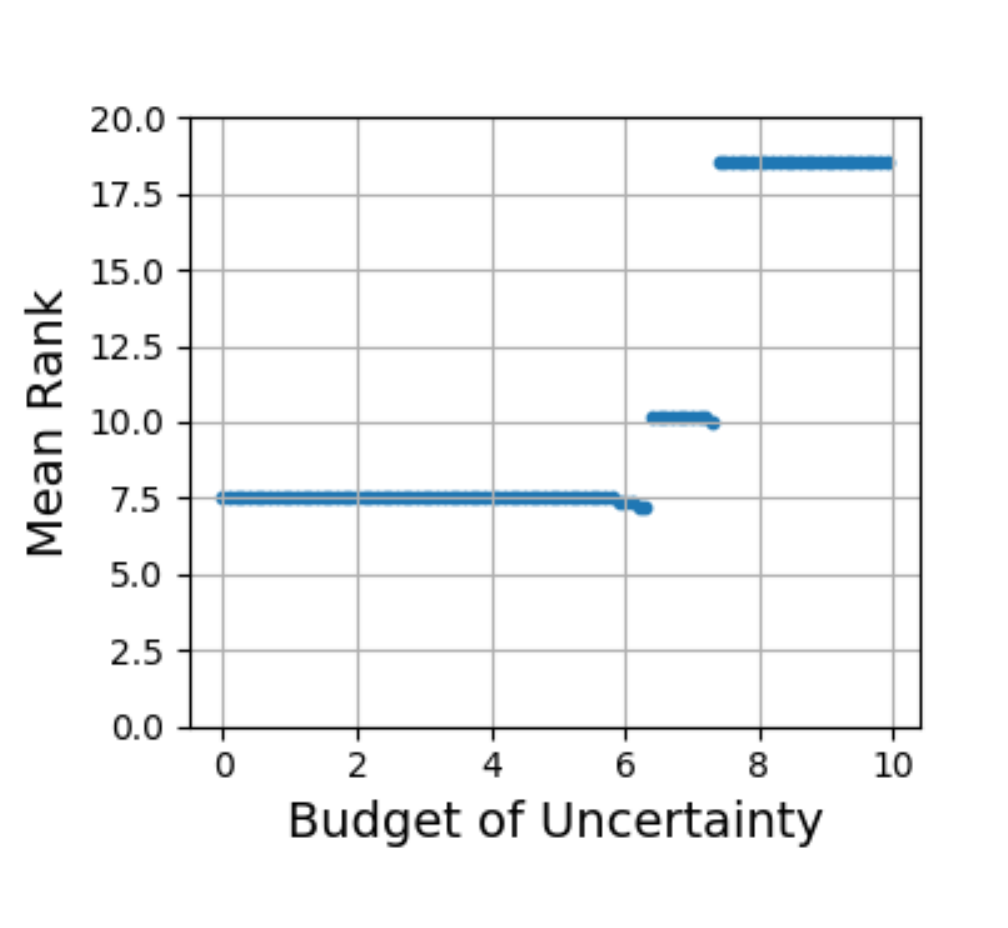}
    \caption{\small \textit{Mean rank} ($M$)}
    \label{fig:exp_results_entropy_rank}
  \end{subfigure}
  \ \ \ 
  \begin{subfigure}[b]{0.23\textwidth}
    \includegraphics[trim= 5 25 25 20, clip,width=0.97\textwidth]{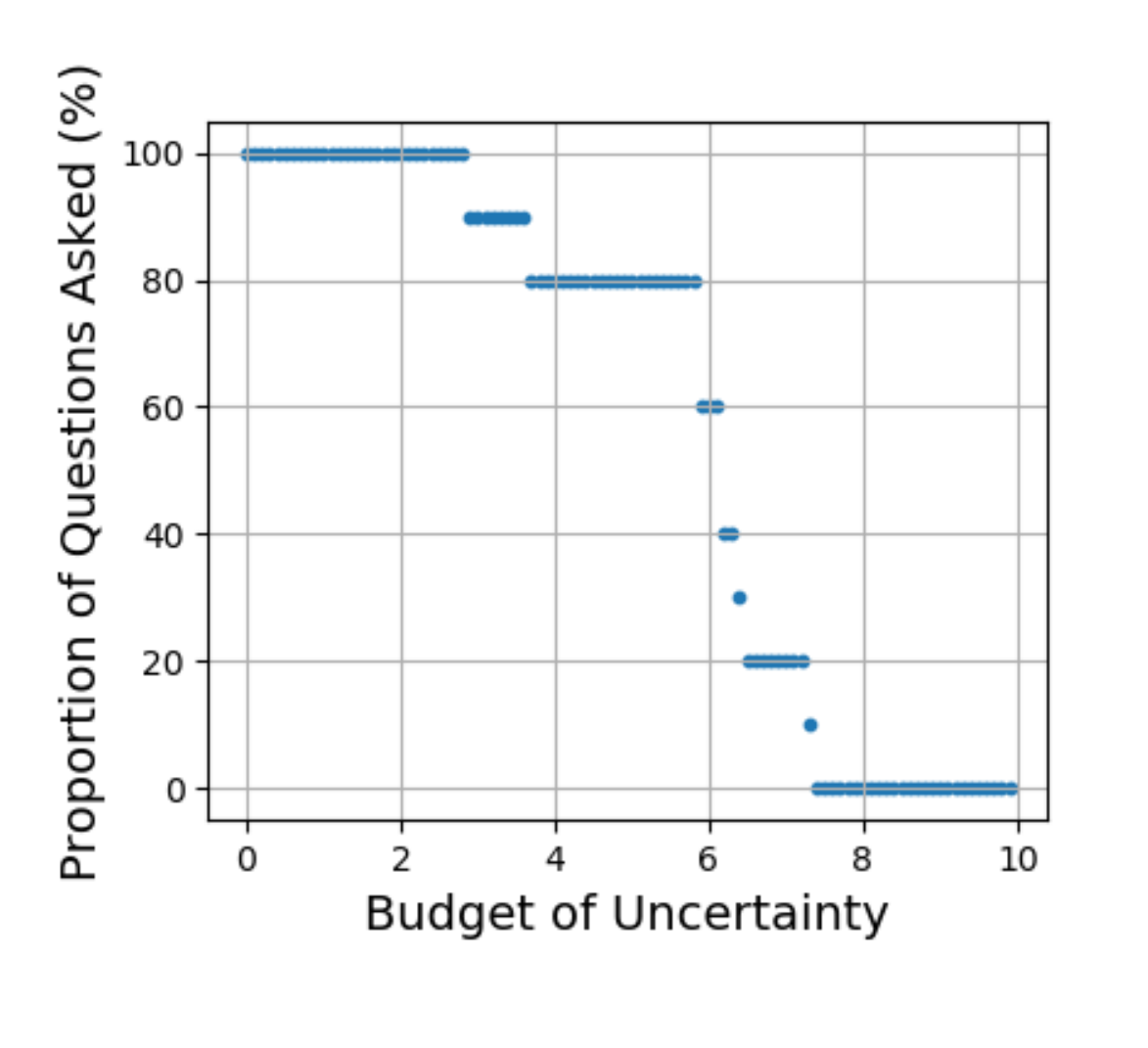}
    \caption{\small Number of queries}
    \label{fig:exp_results_entropy_numQ}
  \end{subfigure}

  \caption{\small Five-step information retrieval results with different \textit{budgets of uncertainty}, using robot-colleceted data.}
  \label{fig:offline_results_qa}
  \vspace{-0.2in}
\end{figure}

Second, we test the uncertainty-driven information retrieval method on our collected dataset. For each POI, the first question from sequence $\mathbf{S}^{'}$ is asked; decisions on whether to ask the next question are based on the uncertainty in the current predicted similarity scores by Pythia-reID. Figure~\ref{fig:exp_results_entropy_rank} shows the \textit{mean rank} achieved for different values of \textit{budget of uncertainty} and Figure~\ref{fig:exp_results_entropy_numQ} shows the corresponding number of queries for the user. The resulting plots appear more discrete than those in Figure~\ref{fig:budget_uncertainty_with_qa} because of the fewer number of search targets in our robotic experiment. Nonetheless, the general trend remains the same and results indicate that the \textit{budget of uncertainty} allows a trade-off between the number of questions that are asked to the user versus the \textit{mean rank} performance of the search algorithm.

\subsection{Online Experiment}

In the online setting, the data is processed serially in the sequence of its arrival. As a consequence, we start with an empty gallery and build incrementally as the video feed is received from the camera. All other parameters are same as the offline case.

For online search, we use a threshold-based maximum-likelihood estimation framework with an infinite time horizon. Our Pythia-reID network is used to first assess the similarity score between the description and the detections from the current image. The closest matching person is chosen from the current gallery based on the similarity score, subject to the minimum score threshold of 0.95. 
Note that even if a match is found, the algorithm continues to improve its hypothesis as additional data arrives; this approach has the benefit of reducing false positives.

\begin{figure}
\centering
  \begin{subfigure}[b]{0.23\textwidth}
    \includegraphics[trim= 5 30 30 30, clip,width=0.95\textwidth]{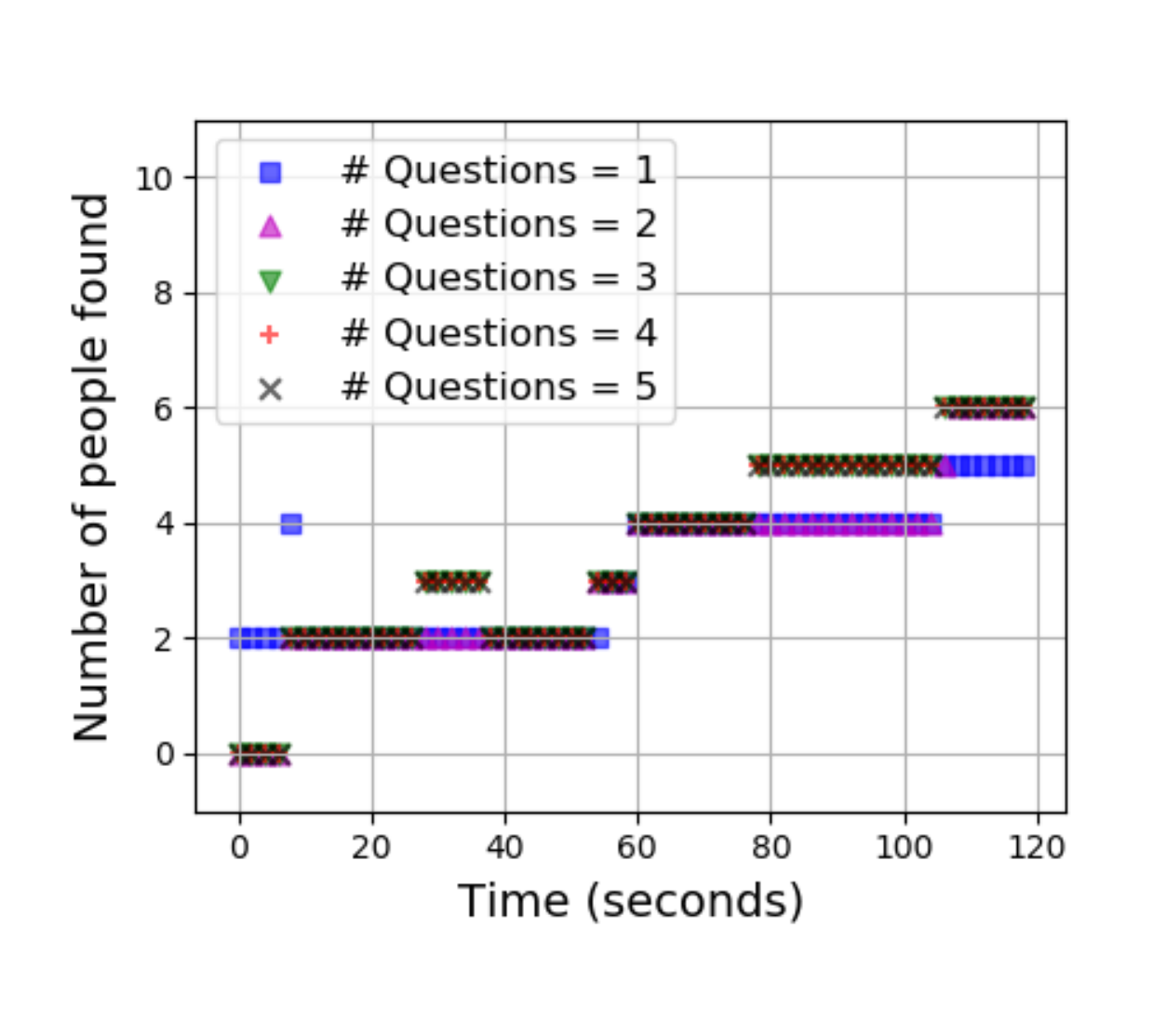}
    \caption{\small Top-1 search results}
    \label{fig:online_results_top1}
  \end{subfigure}
  \ \ \ 
  \begin{subfigure}[b]{0.23\textwidth}
    \includegraphics[trim= 15 30 30 30, clip,width=0.95\textwidth]{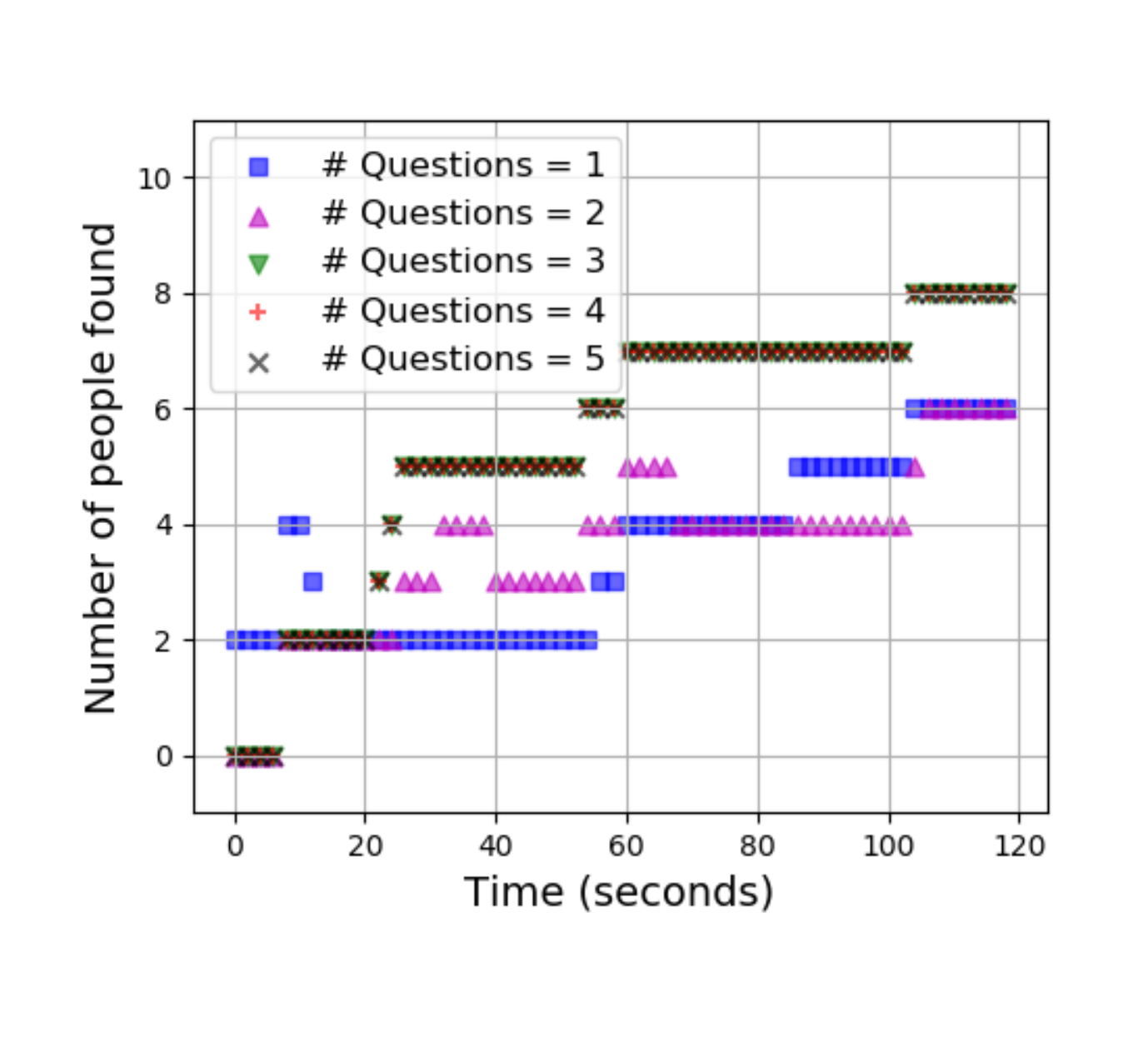}
    \caption{\small Top-10 search results}
    \label{fig:online_results_top10}
  \end{subfigure}

  \caption{\small Online person-search performance, shown over time.}
  \label{fig:online_results}
  \vspace{-0.2in}
\end{figure}

The performance metric used for online case is the number of POIs correctly matched at each frame, as a function of the number of appearance-related questions asked to the user, shown in Figure~\ref{fig:online_results}. Given two independent descriptions for each POI, each can be treated as a separate person without any loss of generality. Thus, effectively, there are 10 POIs in this experiment. Figure~\ref{fig:online_results_top1} and \ref{fig:online_results_top10} shows the number of people found in top-1 and top-10 sense, respectively. One can observe intuitively that more people are found by the search algorithm when higher number of questions are asked to the user. Also, the number of people who are correctly matched rises as time progresses.

\subsection{Discussion}

The offline experiment is relevant when a robot explores the environment first, and the collected video data is available for subsequent search. Since a fixed, a priori gallery of person images is assumed to be known before applying the search algorithm, it is easy to utilize our current model for searching people in the scene. Note that we can also do person-search online, on a robot, in an unconstrained, dynamic environment; examples include rescuing people, handing-over supplies etc. However, to search for a person online, we must add a gating mechanism that only allows matches with similarity scores higher than a set threshold. 

\section{Conclusion}
In this paper, a novel, iterative scheme of obtaining information from the user is developed for performing language-based person re-identification. A human-participant survey was conducted to collect diverse sentence descriptions of person images for evaluating the performance of QA module. An approach to optimize the sequence of questions for faster information collection was developed which can be applied to any other language-based re-ID dataset. Moreover, the uncertainty quantification module enables to regulate the number of questions for the user depending upon uncertainty in the prediction and complements user experience, thus, taking a significant step towards enhancing human-robot interaction in the person search domain. 
The experiments successfully conducted with real-world data from a mobile robot re-enforces our claim that our approach can handle dynamic targets in a crowded environment in both offline and online settings.

\appendices
\section{Submodularity Proof}
\label{sec:appendix}
Consider a gallery of $n$ images $\mathbf{G} = \{g_1, \hdots, g_n \}$ and the set of $n_Q$ questions $\mathbf{Q} = \{q_1, \hdots, q_{n_Q} \}$, asking about person's appearance. $n_Q$ can be arbitrarily large, but finite. Assume that $\mathbf{Q}_A$ and $\mathbf{Q}_B$ are two question sets, satisfying $\mathbf{Q}_A \subseteq \mathbf{Q}_B \subseteq \mathbf{Q}$. For a person-image $i$, responding to $\mathbf{Q}_A$ and $\mathbf{Q}_B$ yields two description sets $\mathbf{d}_{i,A}$ and $\mathbf{d}_{i,B}$, respectively. Denote the description corresponding to $\mathbf{Q}_A$ and $\mathbf{Q}_B$, for the entire gallery as $\mathbf{D}_A = \{\mathbf{d}_{i,A}\}$ and $\mathbf{D}_B = \{\mathbf{d}_{i,B}\}, \ \forall i \in \{1, \hdots,n \}$. 

Based on descriptions $\mathbf{d}_{i,A}$, an \textit{ideal} person search module selects a subset of the gallery $\mathbf{g}_{i,A} \subseteq \mathbf{G}$, satisfying the appearance criteria. Thus, corresponding to $\mathbf{D}_A$ and $\mathbf{D}_B$, we get two sets $\mathbf{G}_A  = \{\mathbf{g}_{i,A}\}$ and $\mathbf{G}_B = \{\mathbf{g}_{i,B}\} \ \forall i \in \{1, \hdots,n \}$.
Since, sentences in $\mathbf{D}_B$ describe people in greater detail than $\mathbf{D}_A$, it should be obvious that for any $i$:
\begin{equation}
    \mathbf{g}_{i,B} \subseteq \mathbf{g}_{i,A}
\label{eq:subset}
\end{equation}
Given the descriptions $\mathbf{d}_{i,A}$, all output images in $\mathbf{g}_{i,A}$ are equally likely to correspond to image $i$. Thus, based on the property of an \textit{ideal} search module, the \textit{rank} is $\frac{|\mathbf{g}_{i,A}| + 1}{2}$, resulting in \textit{mean rank}:
\begin{equation}
    M(\mathbf{D}_A, \mathbf{G}) = \frac{\sum_{i=1}^{n}\frac{\Big( |\mathbf{g}_{i,A}| + 1 \Big)}{2}}{n} =\frac{ \sum_{i=1}^{n}\Big( |\mathbf{g}_{i,A}| + 1 \Big)}{2n}
    \label{eq:exp_rank}
\end{equation}
where, $|\mathbf{g}_{i,A}|$ denotes the cardinality of the set $\mathbf{g}_{i,A}$. As mentioned in \ref{subsec:greedy_algo}, we transform the \textit{mean rank} $M$ into another metric $R = n - M$. Now, assume that $q_e$ is an appearance related question, such that $q_e \in \mathbf{Q} \backslash \mathbf{Q}_B$. The corresponding description and set of images satisfying those descriptions are denoted by $\mathbf{D}_e$ and $\mathbf{G}_e = \{ \mathbf{g}_{i,e}\} \ \forall i \in \{1, \hdots, n\}$, respectively. Define $\Delta$ as the change in performance $R$ by adding the descriptions corresponding to $q_e$, thus:
\begin{equation}
\begin{split}
    \Delta(q_e|\mathbf{Q}_A) & = \Big(n - M(\mathbf{D}_A \cup \mathbf{D}_e, \mathbf{G} )\Big) - \Big(n -  M(\mathbf{D}_A, \mathbf{G}) \Big)\\
    & = M(\mathbf{D}_A, \mathbf{G}) - M(\mathbf{D}_A \cup \mathbf{D}_e, \mathbf{G})\\
    & = \frac{\sum_{i=1}^{n}\Big( |\mathbf{g}_{i,A}| + 1 \Big)}{2n} - 
    \frac{\sum_{i=1}^{n}\Big( |\mathbf{g}_{i,A} \cap \mathbf{g}_{i,e}| + 1 \Big)}{2n}\\
    & = \frac{1}{2n} \sum_{i=1}^{n} \Big( |\mathbf{g}_{i,A}| -  |\mathbf{g}_{i,A} \cap \mathbf{g}_{i,e}| \Big)\\
    & = \frac{1}{2n} \sum_{i=1}^{n} \Big( |\mathbf{g}_{i,A} \cup \mathbf{g}_{i,e}| -  |\mathbf{g}_{i,e}| \Big) 
\end{split}
\end{equation}

Similarly, for $\mathbf{Q}_B$:
\begin{equation}
\begin{split}
    \Delta(q_e|\mathbf{Q}_B) & = \frac{1}{2n} \sum_{i=1}^{n} \Big( |\mathbf{g}_{i,B} \cup \mathbf{g}_{i,e}| -  |\mathbf{g}_{i,e}| \Big) 
\end{split}
\end{equation}

Based on Equation \ref{eq:subset} and from set property, we infer that:
\begin{equation}
\begin{split}
\Big( \mathbf{g}_{i,B} \cup \mathbf{g}_{i,e} \Big) \subseteq \Big( \mathbf{g}_{i,A} \cup \mathbf{g}_{i,e}\Big)
 \implies & |\mathbf{g}_{i,B} \cup \mathbf{g}_{i,e}| \leq |\mathbf{g}_{i,A} \cup \mathbf{g}_{i,e}|\\
 \implies & \Delta(q_e|\mathbf{Q}_B) \leq \Delta(q_e|\mathbf{Q}_A) 
\end{split}
\end{equation}
Hence, we can conclude that for an \textit{ideal} person search module, the performance metric $R$ is submodular.

\bibliographystyle{IEEEtran} 
\bibliography{egbib} 

\end{document}